\documentclass[lettersize,journal]{IEEEtran}
\usepackage{amsmath,amsfonts}
\usepackage{algorithmic}
\usepackage{algorithm}
\usepackage{array}
\usepackage[caption=false,font=normalsize,labelfont=sf,textfont=sf]{subfig}
\usepackage{textcomp}
\usepackage{stfloats}
\usepackage{url}
\usepackage{verbatim}
\usepackage{graphicx}
\usepackage{cite}
\usepackage{xcolor}
\usepackage{multirow}
\usepackage[T1]{fontenc}
\usepackage[colorlinks, 
            linkcolor={blue!70!black}, % 图表、公式链接：深一点的蓝色，更稳重
            citecolor={blue!50!cyan},  % 参考文献引用：淡蓝色/青蓝色
            urlcolor=cyan              % 网页链接：青色
           ]{hyperref}
\usepackage{caption}
\begin{document}

\title{FakeVLM-R1: Internalizing Physical Laws via CoT for Synthetic Image Detection}

% \thanks{Manuscript received January 3, 2026; revised August 16, 2021.}}
% 还要加谁？
\author{
    Leqi Zhu\textsuperscript{\rm 1,2}\textsuperscript{*},
    Junyan Ye\textsuperscript{\rm 1,3}\textsuperscript{*}, 
    Kaiqing Lin\textsuperscript{\rm 4}, 
    Zhiyuan Yan\textsuperscript{\rm 5}, \\
    Conghui He\textsuperscript{\rm 1},
    Weijia Li\textsuperscript{\rm 1,6}\textsuperscript{\dag} 
    \\
    \textsuperscript{\rm 1}Shanghai AI Lab,
    \textsuperscript{\rm 2}Nanjing University,
    \textsuperscript{\rm 3}Sun Yat-Sen University, \\
    \textsuperscript{\rm 4}Shenzhen University, 
    \textsuperscript{\rm 5}Peking University, 
    \textsuperscript{\rm 6}Tsinghua University,
    \thanks{
        \textsuperscript{*}Equal contribution. \quad 
        $^\dagger$Corresponding author.
    }
    % \thanks{Manuscript received: April 20, 2026; revised: }
}

% \markboth{Journal of \LaTeX\ Class Files,~Vol.~14, No.~8, April~2026}%
% \markboth{Submitted to IEEE Transactions on Pattern Analysis and Machine Intelligence, ~2026}%


\maketitle

\begin{abstract}
The development of generative artificial intelligence technologies has propelled the visual realism of synthetic images to an unprecedented level. Although current interpretable detection methods based on Large Multimodal Models (LMMs) have made certain progress, they still rely on imitation learning derived from massive volumes of forged data. Consequently, they lack genuine causal reasoning capabilities and are prone to explanatory hallucinations. To overcome this bottleneck, we propose FakeVLM-R1, aiming to endow the model with human-like critical thinking capabilities when performing synthetic detection tasks. Building upon Supervised Fine-Tuning (SFT), this framework integrates Group Relative Policy Optimization (GRPO) with a Critical Thinking Chain-of-Thought (CoT) mechanism. During the inference phase, the model executes a "bidirectional dialectical reasoning" process: while proposing a forgery hypothesis, it must simultaneously invoke physical commonsense to construct an authenticity counter-proof. Furthermore, we constructed the FakeClue++ dataset with high-quality samples, which extensively introduces annotations guided by the physical laws of authentic images, providing a unified authenticity anchor for the model. Experiments confirm that FakeVLM-R1 achieves SOTA performance the evaluated models across multiple benchmarks. It not only achieves high-precision, logically interpretable detection but also resolves the over-rejection bias of existing methods against real images, demonstrating generalization and robustness against perturbations.
\end{abstract}


\section{Introduction}

\begin{figure}[!t]
\centering
\includegraphics[width=1.0\columnwidth]{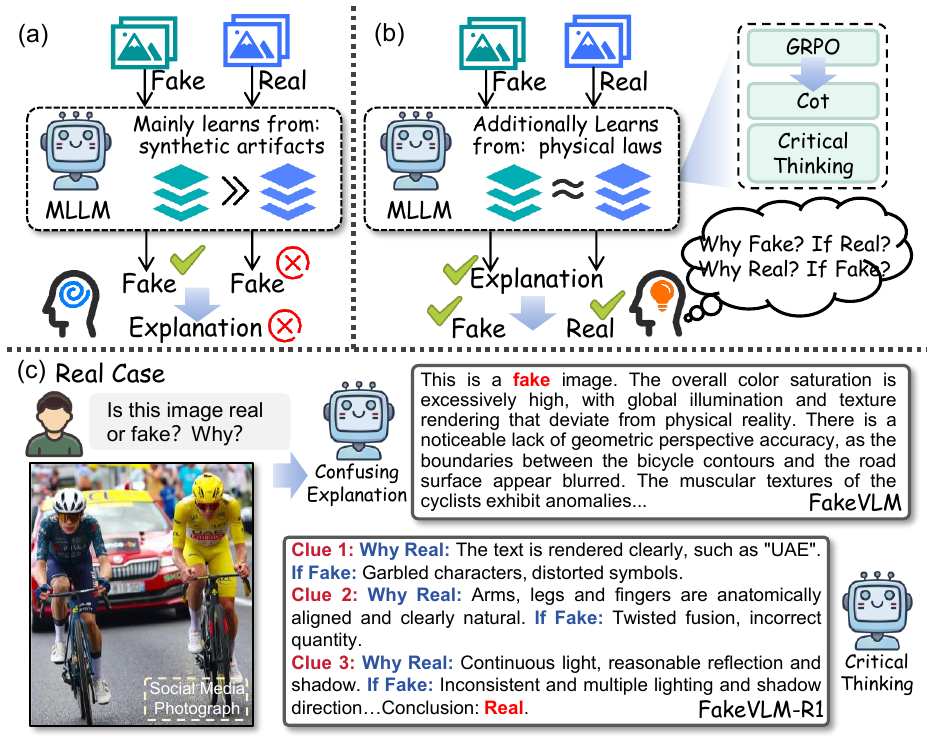}
\caption{Comparison of synthetic detection paradigms. (a) Traditional methods overfit synthetic artifacts, risking explanatory hallucinations; (b) Our FakeVLM-R1 additionally learns real-world physical laws, achieving forensic reasoning via critical thinking; (c) A real-image case demonstrating our method's robustness against false positives.}
\label{comparison}
\end{figure}

\IEEEPARstart{W}{ith} the rapid advancement of multimodal AI-Generated Content (AIGC) technologies, like GANs~\cite{goodfellow2014generative}, Diffusion~\cite{ho2020denoisingdiffusionprobabilisticmodels}, DALL-E~\cite{betker2023improving}, and Z-Image~\cite{imageteam2025zimageefficientimagegeneration}, the visual quality of synthetic images has reached an unprecedented level of photorealism~\cite{dhariwal2021diffusion, rombach2022high, croitoru2023diffusion, ye2025realgen}. While such generative capabilities foster the prosperity of the creative industry, the rise of synthetic data like fake news~\cite{liu2024mmfakebench} or facial forgeries~\cite{mustak2023deepfakes}, simultaneously provokes severe concerns regarding the dissemination of misinformation, fraud, copyright violations, and a broader crisis of public trust~\cite{koutlis2024leveraging, lin2025seeing, jiang2025ivy}. Consequently, developing universal and reliable synthetic detection techniques has emerged as an urgent imperative in the fields of computer vision and multimedia security~\cite{guo2025omniaid}.

Synthetic image detection typically discriminates authenticity by capturing visual artifacts and perceptual defects within the image~\cite{he2021forgerynet, zhu2023genimage, chen2024drct, hong2025wildfake}. Early synthetic detection methods primarily formulated this task as a data-driven binary classification, typically relying on CNNs or Transformer-based visual feature extractors to capture artifactual features associated with spatial pixel anomalies or frequency-domain irregularities~\cite{jeong2022frepgan, chen2024single, li2025fakescopelargemultimodalexpert}. However, these methods suffer from limited transparency and generalization capabilities across diverse content moderation scenarios, and unable to provide convincing evidence in forensic~\cite{kang2025legion}. As the advancements in LMMs enable the exploration of interpretable detection paradigms~\cite{jia2024can}. Data construction studies like LOKI~\cite{ye2024loki} and FakeBench~\cite{li2025fakebench} go beyond merely predicting labels by leveraging natural language to provide detailed logical descriptions of forgery features. Works like X2-DFD~\cite{chen2024x2}, Forensic-Chat~\cite{lin2025seeing}, and AIGI-Holmes~\cite{zhou2025aigi} investigate architectural modifications and novel training strategies to boost the capacity for detection, localization, and explainability. Simultaneously, this explainable forensic process intrinsically bolsters the model's real-vs-fake discriminative ability~\cite{wen2025spot}.

As the preliminary version of this work, FakeVLM~\cite{wen2025spot} has achieved significant advancements in perceptual-level explanation generation. By employing an automated LMMs annotation strategy, it established FakeClue dataset and trained a specialized model for detection and explanation. FakeVLM demonstrated superior performance in general synthetic image detection, exhibiting the ability to precisely localize visual artifacts such as texture distortions and structural anomalies. However, FakeVLM and existing mainstream detection methods remain confined to the traditional Supervised Fine-Tuning (SFT) paradigm as shown in Fig.~\ref{comparison}, facing the following challenges:

\textit{1) Limitations of Imitation Learning:} While SFT endows the model with robust pattern recognition capabilities, which effectively addresses the perceptual challenge of spotting artifacts, however it fundamentally remains a probabilistic mapping rooted in imitation learning, devoid of genuine logical thinking and causal reasoning capabilities~\cite{wang2501critique}. Meanwhile, constrained by this imitative nature of instruction following, the model inherits the defect of explanatory hallucination, where the 'conclusion-first, explanation-subsequent' mechanism frequently leads to factual biases.

\textit{2) Limitations of Datasets:} Model training relies on massive artifact and localization annotations of synthetic images, overlooking the incorporation of real-world physical knowledge, which causes a high false positive rate on real-world images. The model inherently struggles with dialectical reasoning, mechanically misinterpreting natural noise in authentic images as traces of synthesis, revealing a lack of deep-seated confidence in discerning image authenticity.

\begin{table*}[t]
    \centering
    \renewcommand{\arraystretch}{1.2}
    \caption{Comparison and advantages of the proposed FakeVLM-R1 over the preliminary FakeVLM.}
    \label{tab:fakevlm_comparison}
    \vspace{-5pt}
    
    \resizebox{\linewidth}{!}{
        \begin{tabular}{p{3.2cm} p{3.8cm} p{4.2cm} p{7.2cm}}
            \hline % 顶部黑线
            
            % --- 复合表头 ---
            % 使用 multirow 跨两行，并用 multicolumn 保持居中
            \multicolumn{1}{c}{\multirow{2}{*}{Aspect}} & \multicolumn{2}{c}{Model Specification} & \multicolumn{1}{c}{\multirow{2}{*}{Core Advantages}} \\
            
            % 只在第2列和第3列下面画横线
            \cline{2-3} 
            \noalign{\vspace{1pt}}
            
            % 第2行子标题 (第1列和第4列留空，因为上面已经跨行了)
             & FakeVLM \textit{(Preliminary)} & FakeVLM-R1 \textit{(Proposed)} & \\
            
            \hline % 表头下方的黑线
            \noalign{\vspace{3pt}}
            
            \textbf{Training Paradigm} & SFT & SFT + GRPO & Suppresses Hallucinations: Shifts from passive imitation to active exploration. \\
            \noalign{\vspace{3pt}}
            \hline
            \noalign{\vspace{3pt}}
            \textbf{Reasoning Style} & Unidirectional explanation & Bidirectional dialectical CoT & Forensic-level Logic: Enforces self-consistency via counter-proofs, avoiding conclusion-first bias. \\
            \noalign{\vspace{3pt}}
            \hline
            \noalign{\vspace{3pt}}
            \textbf{Dataset Base} & FakeClue (100K samples) & FakeClue++ (50K samples) & High Efficiency \& Low Cost: Achieves better generalization with half the data by annotating structured physical laws. \\
            \noalign{\vspace{3pt}}
            \hline
            \noalign{\vspace{3pt}}
            \textbf{Real-image Priors} & Limited amount & Explicitly annotated & Eliminates Over-rejection: Establishes authenticity anchors to prevent misclassifying real noise as forgeries. \\
            \noalign{\vspace{3pt}}
            
            \hline % 底部黑线
        \end{tabular}
    }
    \vspace{-5pt}
\end{table*}

To transcend these limitations, this study conducts a systematic expansion at both the methodological and data levels as shown in Table~\ref{comparison}, proposing FakeVLM-R1—an advanced iteration designed to orchestrate the evolution from perceptual detection to logical reasoning. At the algorithmic level, we integrate a Reinforcement Learning (RL) training framework and a Critical Thinking Chain-of-Thought (CoT)~\cite{wei2022chain} mechanism atop the foundation of SFT, utilizing Group Relative Policy Optimization (GRPO) to train the model's bidirectional dialectical reasoning capabilities. This reasoning paradigm necessitates that the model engage in explicit dialectical scrutiny before generating a conclusion: while proposing a 'forgery hypothesis' based on artifact clues, it must simultaneously invoke physical knowledge to construct an authenticity counter-proof (e.g., questioning whether a specific blur stems from a generative defect or results from optical physics like large-aperture depth of field). Ultimately, by validating judgments through this rigorous logical self-consistency, the model significantly suppresses the hallucination issues prevalent in detection tasks while concurrently reducing the false positive rate on authentic images. At the data level, we executed a strategic overhaul of our data foundation, establishing FakeClue++, a dataset comprising 50k samples. FakeClue++ enables precise region-level localization and deep logical explication of artifacts. More crucially, to address the critical deficiency regarding the model's inability to comprehend authenticity, we have meticulously curated a reasoning-oriented dataset of real imagery, providing granular descriptions of physical plausibility in authentic photography. This enables the SFT process to simultaneously assimilate knowledge from two dimensions: on one hand, it leverages meticulously expert-annotated synthetic fake imagery to sharpen the model's sensitivity to subtle artifact features; on the other hand, by learning from extensive real-world imagery, it compels the model to internalize objective physical world knowledge, such as lighting consistency and perspective relations. Compared to exhaustively enumerating ever-changing forgery traces, describing the inherent physical laws of authentic images is significantly more straightforward and structured. This relatively simple and highly efficient annotation paradigm not only drastically reduces the cost of dataset construction but also provides the model with a unified and robust authenticity anchor. FakeClue++ effectively bridges the cognitive gap in reasoning depth present in our previous work, enabling the model to achieve superior generalization with a more compact data volume.

In summary, the main contributions of this work are as follows:
\begin{itemize}
\item We propose FakeVLM-R1, a novel framework designed to propel synthetic image detection from visual perception matching to forensic-level logical reasoning. By integrating GRPO reinforcement learning with an explicit CoT, the model is endowed with robust bidirectional dialectical reasoning capabilities, fundamentally suppressing explanatory hallucinations and high false positive rates.
\item We construct FakeClue++, a high-quality interpretable synthetic detection dataset. Beyond incorporating expert-level, fine-grained artifact annotations, it innovatively introduces extensive real-image annotations guided by intrinsic physical laws, providing the model with a unified and robust authenticity anchor at exceptionally high data efficiency.
\item Extensive experiments demonstrate that FakeVLM-R1 achieves state-of-the-art performance across multiple mainstream benchmarks. It not only significantly outperforms existing LMMs and interpretable baselines in overall detection accuracy but also achieves a paradigm-level leap in unseen-domain generalization, robustness against perturbations, and forensic-level reasoning depth.
\end{itemize}
\section{Related works}
Recently, the evolution of LMMs has opened new avenues for transitioning from traditional black-box discrimination to interpretable forensics in synthetic detection. In this section, we first outline the progression of data-driven synthetic detection techniques, followed by an overview of current developments and existing limitations in explainable synthetic detection.

\subsection{Data-driven synthetic detection}
With generative models excelling in various tasks~\cite{ye2025echo, yan2025gpt, he2026mind}, discriminating the realism of synthetic images has become essential. Traditional synthetic image detection was primarily formulated as a data-driven binary classification task~\cite{wang2020cnn, ojha2023towards, bi2023detecting}. Various deep neural networks were employed to capture the specific visual artifacts and feature deviations left by generative models during the synthesis process~\cite{goodfellow2020generative}. Depending on the forensic cues utilized, current methods fall into two main categories: flaw-based~\cite{zhong2023rich, ciftci2020fakecatcher} (e.g., texture or physical rules) and fingerprints-based~\cite{wolter2022wavelet, wang2020cnn} (e.g., imperceptible patterns like frequency artifacts). In the spatial domain, extensive research employs CNN to extract pixel-level generative traces, such as checkerboard artifacts, color bleeding, and local texture over-smoothing or inconsistencies~\cite{tan2024rethinking, gupta2024visual, wani2025comparative, lađević2024detection}. Recently, ViT have been introduced for global feature modeling to capture deep semantic contradictions resulting from the limited receptive fields of generative models, including conflicting lighting directions, distorted perspective geometry, and long-range texture distortions~\cite{hossain2023advancing, chang2023antifakeprompt}. In the frequency domain, addressing the inherent physical defects of upsampling operations in generators, research focuses on extracting periodic high-frequency noise, abnormal energy distribution spikes, and phase spectrum distortions from spectrograms~\cite{liu2022fad, tan2024frequency, wang2023dynamic}. However, despite their superior performance on specific datasets, these methods fundamentally remain black-box models based on low-level statistical feature mapping. They output only a single authenticity probability score, lacking the integration of high-level logical semantics and explainability.

\subsection{Explainable synthetic detection}
With the rapid proliferation of LMMs, advanced foundation models (e.g., GPT~\cite{gpt4o}, Gemini~\cite{google2025gemini}, LLaVA~\cite{liu2023visual}, Qwen-VL~\cite{bai2025qwen25vltechnicalreport}) have demonstrated robust vision-language alignment and logical articulation capabilities, which empowers the models to output explanations while performing binary classification~\cite{wu2023q, zhang2024visually, jia2024can}. However, since general-purpose LMMs are primarily pre-trained on broad internet image-text data, they struggle to directly undertake complex forensic tasks targeting fine-grained forgery clues. Therefore, extensive research has focused on the synthetic data level to construct universal explainable datasets~\cite{ye2024loki, li2025fakebench, kang2025legion}. Furthermore, within the context of synthetic image detection tasks, recent works have further explored the capabilities of LLMs in artifact explanation and forgery localization~\cite{xu2024fakeshield, zhang2024common, huang2024ffaa}. On this basis, strategies like adopting hybrid expert models~\cite{jia2024can, chen2024x2} or enhancing the training paradigm~\cite{wen2025busterx++, jiang2025ivy, li2025skyra} have emerged as pivotal factors in realizing highly accurate detection. These works range from general synthetic images, facial forgeries, and local manipulations to social media hybrid forgeries and spatiotemporal video artifacts~\cite{liu2024forgerygpt, huang2025sofake, wen2025spot, huang2025sida}, realizing an all-in-one professional forensic model that integrates 'detection, localization, and explanation'. Works like FakeVLM~\cite{wen2025spot} successfully achieves precise localization and natural language explication for various perceptual and dynamic artifacts, such as abnormal texture over-smoothing, abrupt edge mutations from splicing, and spatiotemporal logical collapses. However, they heavily rely on feature learning from large-scale forged images. Their data construction overly emphasizes 'describing forgery traces' while neglecting the internalization of 'intrinsic physical laws', leading to extremely high false positive rates when models are confronted with complex real-world scenarios. These critical pain points serve as the core motivation for introducing reinforcement learning and bidirectional dialectical reasoning in this study.
\begin{figure*}[!t]
\centering
\includegraphics[width=2.0\columnwidth]{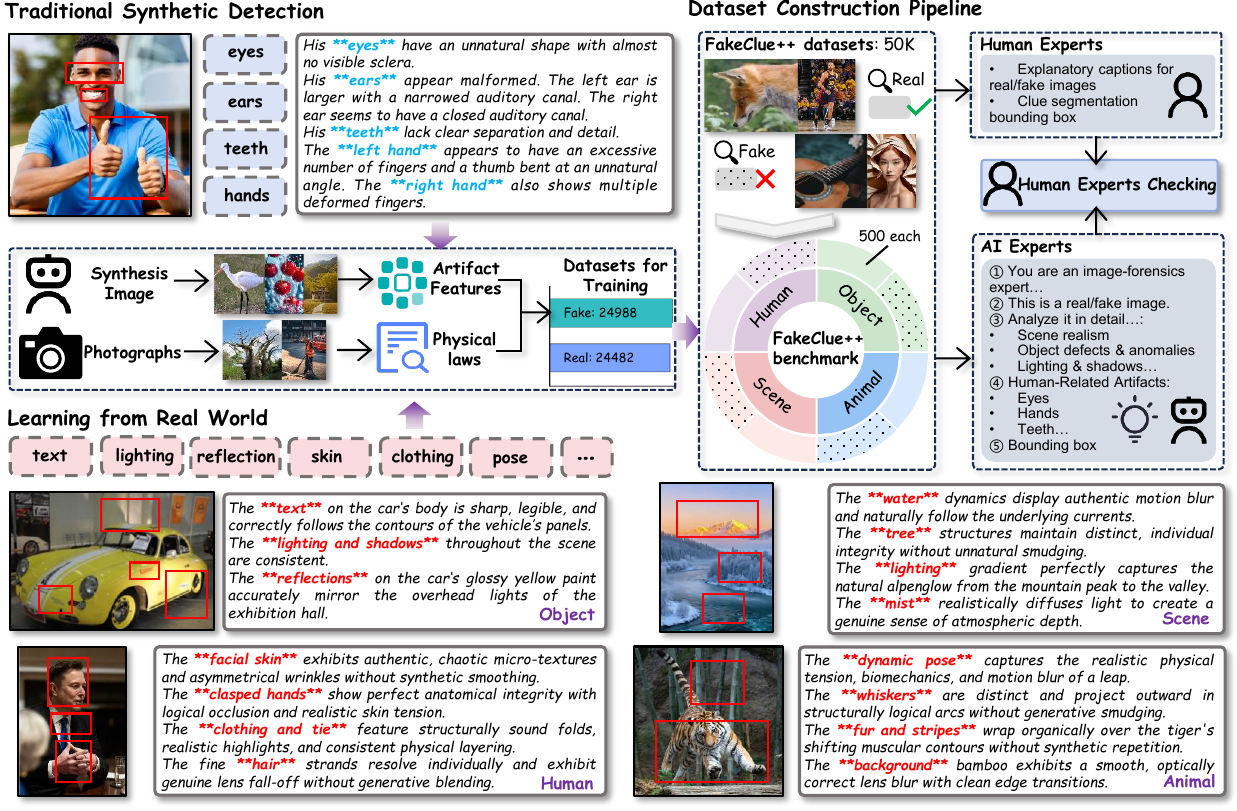}
\caption{Overview of the FakeClue++ dataset. Diverging from traditional datasets focused on artifact annotation, FakeClue++ additionally incorporates the internalization of real-world physical laws. By incorporating a vast amount of authentic images that inherently conform to natural rules, it enables highly efficient and low-cost annotation via large models.}
\label{dataset}
\end{figure*}

\section{Dataset}
FakeClue++ is developed as an advanced iteration of the existing interpretable dataset FakeClue, with the objective of enabling the model to simultaneously acquire fine-grained forgery feature representations and internalize physical world knowledge, as shown in Fig.~\ref{dataset}. In this section, we detail the dataset construction pipeline, encompassing data collection, preprocessing, and annotation. Furthermore, we highlight the key distinctions between FakeClue++ and existing datasets.
\subsection{Data Collection}
To guarantee data diversity and comprehensive coverage, we curated a core subset of 50,000 challenging samples, comprising both typical forgeries and authentic images, from multiple heterogeneous sources including FakeClue~\cite{wen2025spot}, SynthScars~\cite{kang2025legion}, OpenImage~\cite{Kuznetsova_2020}, and Internet(photography platforms and social media).

\textit{1) Fake Image:} FakeClue serves as the foundational baseline, recognized as the pioneer large-scale synthetic detection dataset leveraging LMMs for automated artifact description. While it encapsulates rich generative semantics, its original annotations primarily rely on the visual perception of LMMs, lacking fine-grained localization and accuracy verification. To address this, we integrated SynthScars, distinguished for its focus on fully synthetic imagery and expert-annotated fine-grained artifact taxonomy. This integration significantly bolsters the dataset's coverage of complex structural distortions and synthetic anomalies. Furthermore, we supplemented the dataset with AI-generated images sourced from social media platforms.

\textit{2) Real Image:} To transcend mere artifact recognition and facilitate the mastery of intrinsic physical laws (e.g., lighting consistency, specular reflection, material texture, and optical depth of field), we incorporate the learning of Physical Priors of Authenticity. To this end, we curated high-quality authentic samples from OpenImages and extensively crawled real-world photography from social media platforms. These authentic images serve as positive anchors, compelling the model to contrast physically consistent reality against counter-factual forgeries, thereby sharpening its sensitivity to violations of physical common sense. Such strategy of learning from authentic samples provides a novel perspective for synthetic detection.

\subsection{Data Preprocess}
To ensure data reliability and precise task definition, we implemented a rigorous data cleaning and multi-dimensional annotation pipeline prior to the interpretable annotation phase. All collected images underwent multiple rounds of authenticity verification and quality screening to guarantee absolute Ground Truth accuracy. Subsequently, the data was refined into four core categories: Human, Object, Scene, and Animal. This fine-grained categorization aims to accommodate differentiated reasoning requirements governed by distinct counterfeit features and physical laws (e.g., human anatomical structure, animal fur texture, and scene lighting and perspective), thereby providing accurate prior knowledge for subsequent annotations. The processed data was scientifically partitioned into two independent subsets: the majority is utilized for the SFT stage to construct the model's capabilities in precise artifact localization and logical reasoning, while the remaining 4,000 images constitute the standalone FakeClue++ Benchmark, dedicated to evaluating detection robustness, category generalization, and the logical self-consistency of generated explanations on unseen domain samples.

\subsection{Multiple Annotation}
To strike a balance between large-scale annotation efficiency and forensic-level logical depth, we construct a sophisticated annotation pipeline. Initially, for the incorporated SynthScars subset, we directly leverage its high-precision pixel-level masks and fine-grained artifact annotations derived from human experts, providing the dataset with high-confidence negative anchors and robust supervision signals. For broader data sources, We first leverage the advanced Gemini API to act as a forensic expert, generating preliminary authenticity judgments, artifact localizations Bounding Boxes, and descriptions based on predefined prompts across dimensions such as lighting consistency, scene logic, and anatomical plausibility (e.g., facial features and limbs). Subsequently, human experts intervene in a rigorous verification loop, focusing on rectifying potential AI hallucinations and refining the descriptive text from a physical perspective. Ultimately, this pipeline yields a structured Explanatory Check List, which not only encapsulates specific artifact explanations but also enriches annotations of fundamental physical attributes, such as lighting, reflections, and material textures, via expert knowledge, thereby ensuring the model establishes a deep causal mapping from visual representations to physical laws. Notably, annotating real images inherently involves commonsense knowledge, the annotation model can yield highly logical explanations simply by describing its direct visual perceptions.

\subsection{Comparisons with Existing Datasets}
Existing datasets often struggle to strike a balance between quality and scale. 
While traditional synthetic detection datasets like GenImage~\cite{zhu2023genimage} possess massive data volume, they often fall short in achieving cross-modal alignment between vision and language. While datasets like DD-VQA~\cite{zhang2024common}, LOKI~\cite{ye2024loki}, and FakeBench~\cite{li2025fakebench} ensure quality through manual annotation, their limited scale, ranging from 3k to 6k samples, is insufficient for the robust training of large models. The time-consuming and labor-intensive nature of expert-annotated data is inevitable. Conversely, although MMTD-SetA~\cite{xu2024fakeshield} and FakeClue~\cite{wen2025spot} achieve a scale of 100k+, they are constrained by either a narrow focus on single domain tampering (rather than synthesis) or a reliance on automated LMMs annotations, which often lack explanation accuracy check. More critically, learning from authentic-synthetic imbalanced data makes the model prone to internalizing biases toward fake data, leading to overfitting on similar data distributions and a lack of generalizability.

FakeClue++ bridges this gap by achieving an optimal equilibrium. Designed as a benchmark for General Synthesis detection, it adopts a unique collaborative annotation paradigm. This approach constructs 50k high-quality samples while preserving fine-grained categorization. The annotation of FakeClue++ is built upon a small fraction of existing synthetic detection benchmarks, and crucially, we supplement it with large-scale authentic data, achieving annotation through a cost-effective and highly accurate approach. It is worth emphasizing that, although FakeClue++ comprises only half the volume of the original FakeClue, our subsequent experiments compellingly demonstrate that the high density of physical priors and logical reasoning within the data, enables the model to achieve superior generalization and ideal detection performance with significantly reduced data dependency.

\section{METHOD}

\begin{figure*}[!t]
\centering
\includegraphics[width=2.0\columnwidth]{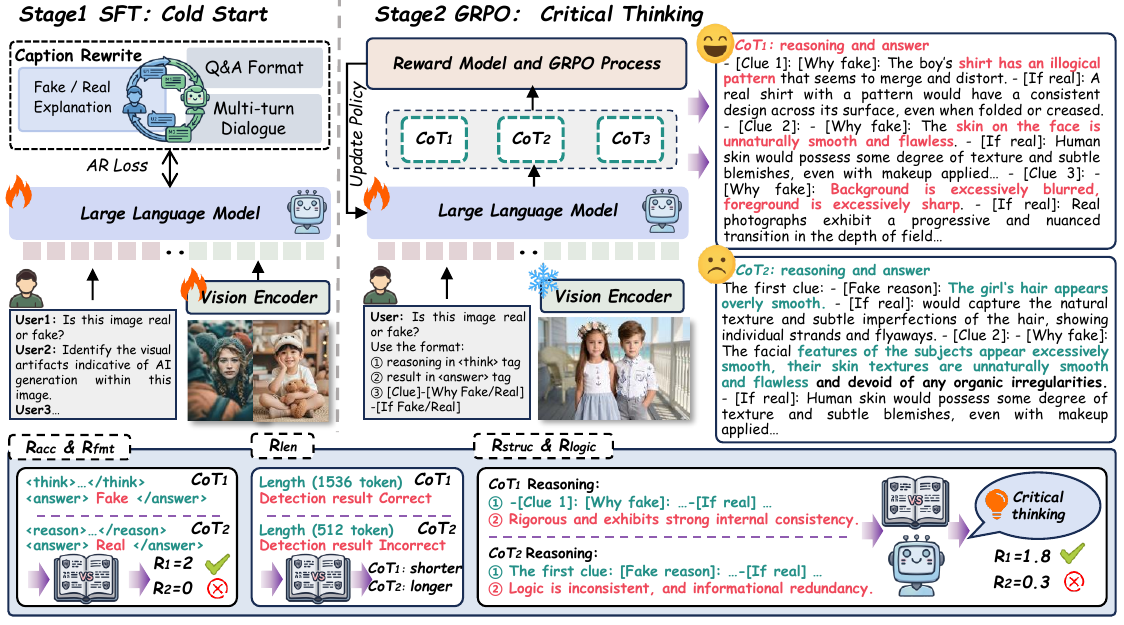}
\caption{The two-stage training framework of FakeVLM-R1. To cold-start the explanatory generation capability, we fine-tune both the Vision Encoder and the LLM Decoder. During RL Stage, we optimize the LLM decoder using multiple rewards to stimulate bidirectional dialectical reasoning and precise detection.}
\label{method}
\end{figure*}

In this section, we introduce FakeVLM-R1 according to Fig.~\ref{method}, which aims to propel the evolution of synthetic image detection from perceptual pattern recognition to active logical reasoning. To endow the model with critical thinking akin to human experts, we design a progressive two-stage training framework. This framework first ensures foundational visual perception through a Cold Start phase. Subsequently, it incorporates GRPO and leverages CoT training to unleash the model's deep dialectical reasoning potential, thereby achieving precise discrimination and the generation of logically self-consistent explanations. As an MLLM post-training framework, FakeVLM-R1 effectively adapts to various architectures, such as the Qwen-VL~\cite{bai2025qwen25vltechnicalreport} or LLaVA~\cite{liu2023visual} series, and enhances practical applicability through deployment on efficient inference frameworks like vLLM~\cite{kwon2023efficient} and SGLang~\cite{zheng2024sglang}.

\subsection{Stage 1: Cold Start of Explainable Detection}
The primary objective of the first stage is to construct a cold start equipped with foundational visual perception capabilities and diverse output formats. Architecturally, we adopt a standard Vision-Language Model structure, where image features extracted by the Vision Encoder are aligned with text prompt embeddings in the latent space and fed into the LLM Decoder. To address the scarcity of high-quality reasoning data, we leverage LLMs to rephrase the interpretable annotations of FakeClue++. This process incorporates diverse QA formats and multi-turn dialogue transformations. We design strategies such as asking the model to describe visual content before analyzing potential artifacts, or alternatively, requiring an initial authenticity judgment followed by an explanation of the rationale. The overall objective of this pipeline is to enhance the model's generalization capability. In this phase, the model undergoes full-parameter SFT by minimizing the auto-regressive loss. While this procedure empowers the model to recognize manifest texture distortions or anatomical anomalies, its underlying nature is inherently that of probabilistic imitation learning, rendering it prone to hallucination.

\subsection{Stage 2: Detection through Critical Thinking}
The second stage serves as the core of this study, aiming to internalize critical thinking via a Reinforcement Learning mechanism. Departing from the traditional unidirectional direct reasoning paradigm (the sequential enumeration of forensic clues.), we design a unique 'bidirectional dialectical' reasoning paradigm. Within the \texttt{<think>} tag, the model is mandated to engage in a dual-perspective self-examination for each visual clue: specifically, while proposing a 'Forgery Hypothesis' (in [Why fake] tag) based on artifact evidence, it must continuously invoke physical world knowledge to construct an 'Authenticity Counter-proof' (in [If real] tag), and vice versa ([Why real] - [If fake]). For instance, when analyzing an image of reflections on vehicle body, the model must weigh whether the reflection is a natural phenomenon consistent with the laws of optical physics or a rendering flaw indicative of a generative model. This mechanism compels the model to undergo a comprehensive logical verification process before generating the final conclusion in the \texttt{<answer>} tag.

In terms of algorithmic optimization, we employ GRPO, an efficient reinforcement learning paradigm that eliminates the need for a traditional Critic Model. For each combination of input image $I$ and prompt $q$, the policy model $\pi_\theta$ samples a group of reasoning paths and responses $\{o_1, o_2, ..., o_G\}$ of size $G$. Subsequently, we introduce a set of multi-dimensional reward functions to comprehensively evaluate this group of outputs. GRPO directly estimates advantage values by calculating the relative mean and standard deviation of rewards within the group. This mechanism leverages intra-group competition to effectively reinforce reasoning patterns that exhibit superior logical depth and adherence to physical commonsense. The optimization objective is to maximize the following objective function:
\begin{equation}
\label{eq:grpo_objective}
\begin{split}
J_{GRPO}(\theta) &= \mathbb{E}_{q \sim P(Q), \{o_i\}_{i=1}^G \sim \pi_{\theta_{old}}} \Bigg[ \frac{1}{G} \sum_{i=1}^G \bigg( \\
&\quad \min \left( \rho_i(\theta) \hat{A}_i, \text{clip}\left(\rho_i(\theta), 1-\epsilon, 1+\epsilon\right) \hat{A}_i \right) \bigg) \Bigg]\\
\end{split}
\end{equation}

where $\rho_i(\theta) = \frac{\pi_\theta(o_i|q)}{\pi_{\theta_{old}}(o_i|q)}$, $\pi_{\theta_{old}}$ denotes the old policy. And we calculate the advantage function $\hat{A}_i$ as follows:

\begin{equation}
\label{eq:advantage_explicit}
\hat{A}_i = \frac{R_i - \mathrm{Mean}(\{R_1, \dots, R_G\})}{\mathrm{Std}(\{R_1, \dots, R_G\}) + \delta}
\end{equation}

\subsection{Reward Functions for GRPO}
To steer the model during the Reinforcement Learning CoT process toward balancing detection accuracy, adherence to reasoning formats, and the depth of dialectical thinking, we design a composite reward system encompassing multiple distinct dimensions. For each sampled response $o_i$, the total reward $r_i$ is derived via a weighted summation of the following five components:

\begin{equation}
\label{eq:total_reward_sum}
r = R_{acc} + R_{fmt} + R_{struc} + R_{logic} + R_{len}
\end{equation}

To guarantee fundamental task compliance, we employ two deterministic rule-based rewards drawing inspiration from DeepSeek-R1~\cite{guo2025deepseek}. The Correctness Reward ($R_{acc}$) serves as a sparse objective, assigning a positive score solely when the verdict extracted from the \texttt{<answer>} tag aligns with the ground truth. Complementing this, the Format Reward ($R_{fmt}$) enforces structural validity via strict regular expressions. The reasoning process is encapsulated within \texttt{<think>} tags, and the \texttt{<answer>} tags are strictly confined to the binary verdict ('Real' or 'Fake'). This ensures the integrity of the CoT process.

To ensure both the structural integrity and semantic rigor of the CoT, we implement two synergistic rewards. First, the Thinking Structure Reward ($R_{struc}$) enforces the proposed 'bidirectional dialectical' paradigm. By utilizing regular expressions, this function assigns positive rewards solely to responses that explicitly pair each visual clue with both an 'Authenticity Counter-proof' and a 'Forgery Hypothesis', while penalizing any degeneration into unidirectional descriptions:  
\begin{equation}
R_{\text{struc}}(o_i) =
\begin{cases}
+1.0 & \text{if } \mathcal{M}(o_i, \mathcal{P}_{\text{dialectic}}) \\
-1.0 & \text{otherwise}
\end{cases}
\end{equation}
where the pattern $\mathcal{P}_{dialectic}$ represents the structural sequence '[Clue]' $\to$ '[Why real/fake]' $\to$ '[If fake/real]'. Complementing this structural constraint, the Logical Consistency Reward ($R_{logic}$) evaluates the intrinsic quality of the reasoning. We employ SophiaVL-R1~\cite{fan2025sophiavl}, a high-capability open-source VLM, as an external Critic to score the logical self-consistency and causal validity of the content (from 0.0 to 1.0). This 'structure-semantic' dual mechanism ensures that the generated explanations are not only formatted correctly but are also logically sound and persuasive.

Moreover, we implement a Cosine Length Reward to dynamically optimize the reasoning budget. Our objective is twofold: efficiency for success and exploration for failure. Specifically, when the model answers correctly, we incentivize conciseness to prevent redundancy. Conversely, when the model errs, we encourage longer reasoning chains, penalizing short or hasty judgments to force deeper cognitive exploration. Formally:

\begin{equation}
\label{eq:len_reward_linear}
R_{len}(o_i) = 
\begin{cases} 
\min(l - L_{max}, 0) & \text{if } y_{pred} \neq y_{gt} \\
\min(-l + L_{max}, 0.5 L_{max}) & \text{if } y_{pred} = y_{gt}
\end{cases}
\end{equation}

where $L_{max}$ represents the maximum reward length, and $l$ is the generated sequence length. The final score is clipped at an upper bound to prevent excessive rewards. 
\section{EXPERIMENTS}
\subsection{Protocols}
In this section, we present a systematic evaluation of the proposed FakeVLM-R1, analyzing its performance across three key dimensions:
\begin{itemize}
    \item We evaluate the model's efficacy in distinguishing between Real and Fake images, and scrutinize its capability as a forensic tool by assessing the quality of the generated interpretable content.
    \item We employ broader datasets and diverse forms of input data. This allows us to rigorously test the model's generalization capabilities in cross-domain scenarios and its robustness against interference.
    \item We validate the effectiveness of distinct modules within the model architecture through ablation experiments. We specifically analyze the critical role of GRPO and CoT in enhancing overall detection.
\end{itemize}
Detailed experimental setups are provided below.

\textit{1) Experimental Benchmarks:} 
This study evaluates on two interpretable benchmarks: the proposed FakeClue++ and LOKI~\cite{ye2024loki}. FakeClue++ is designed by integrating physical priors with an Explanatory Check List, it rigorously assesses a model's ability to capture subtle traces that violate physical common sense. It consists of 4,000 evenly distributed real and fake samples spanning four categories. Complementing this, LOKI is a multimodal benchmark comprising over 2,200 diverse images from mainstream generators (e.g., FLUX, Midjourney, SD). It provides fine-grained explanatory annotations that go significantly beyond traditional binary classification. In addition, we also utilize two other datasets: MMFakeBench~\cite{liu2024mmfakebench} and DMimage~\cite{corvi2023detection}, both of which are synthesis detection datasets based on standard classification paradigms.

\textit{2) Baseline Models:}
We benchmark the proposed method against four categories of mainstream synthetic detection paradigms: (1) data-driven binary classifiers (Effort~\cite{yan2024effort}, NPR~\cite{tan2024rethinking}, AIDE~\cite{yan2024sanity}); (2) closed-source LMMs (GPT-4o~\cite{gpt4o}, GPT-5~\cite{openai_gpt5_system_card_2025}, Gemini-2.5~\cite{google2025gemini}); (3) open-source LMMs (DeepseekVL2~\cite{wu2024deepseekvl2mixtureofexpertsvisionlanguagemodels}, Intern3-VL~\cite{zhu2025internvl3exploringadvancedtraining}, Qwen2.5-VL~\cite{bai2025qwen25vltechnicalreport}); and (4) interpretable detection LMMs (Skyra~\cite{li2025skyra}, SIDA~\cite{huang2025sida}, BusterX++~\cite{wen2025busterx++}, FakeVLM~\cite{wen2025spot}). All baseline models utilize their official pre-trained weights, with input and output formats minimally adapted to align with specific tasks and evaluation protocols.

\textit{3) Evaluation Metrics:} We adopt a dual evaluation framework comprising quantitative detection performance and qualitative interpretability. In terms of detection, beyond standard metrics such as Total Accuracy and F1-score, we emphasize the independent analysis of Real/Fake class-wise accuracy and fine-grained sub-category metrics. This configuration aims to rigorously verify whether the model balances keen sensitivity to synthetic artifacts with a robust understanding of real-world physical laws, thereby preventing class bias. Regarding interpretability, we employ Gemini-2.5-Pro as an automated LMM-as-a-judge to compare model explanations according to a checklist, which was meticulously constructed through a human-in-the-loop agentic workflow. The evaluator is prompted without access to model identities, and it quantitatively scores explanatory texts across three dimensions: Relevance (consistency with visual content), Logicality (coherent reasoning without redundancy), and Completeness (coverage of details and core information), to comprehensively assess the quality of reasoning output.

\textit{4) Implementation Details:}
Implemented within the PyTorch framework, all training and evaluation procedures were executed on a cluster of 8 NVIDIA H200 141GB GPUs. FakeVLM-R1 utilizes Qwen2.5-VL as its backbone. During the SFT stage, the model undergoes full-parameter fine-tuning for one epoch. In the subsequent GRPO stage, the vision encoder is frozen, and only the LLM decoder is fine-tuned for one epoch. Furthermore, we enforced strict data isolation protocols for experiments to prevent test set leakage, specifically, we meticulously ensured that the training data of FakeClue++ has absolutely no overlap with any of the evaluation benchmarks, guaranteeing rigorous data separation. For LOKI-based benchmarks, we adopted two distinct evaluation configurations: a direct questioning format and the original VQA paradigm. To ensure a fair comparison, we standardize basic experimental settings, such as random seeds.

\begin{table*}[t]
    \centering
    \renewcommand{\arraystretch}{1.0}
    \caption{Detection accuracy and explainability metrics evaluated on FakeClue++. All results are expressed as percentages (\%). \textbf{Bold} and \underline{underlined} values indicate the best and second-best results, respectively.}
    \label{tab:fakeclue++_eval}
    \vspace{-5pt}
    % 在 table* 环境下，\linewidth 指的是整个页面的宽度（即两栏的总宽），
    % 所以 resizebox 会自动把表格拉宽到通栏大小。
    \resizebox{\linewidth}{!}{
        \begin{tabular}{lccccccccccc}
            \hline % 顶部黑线
            
            % --- 复合表头 ---
            \multicolumn{1}{c}{Method} & \multicolumn{4}{c}{Total Detection} & \multicolumn{4}{c}{Type-Specific (Acc)} & \multicolumn{3}{c}{Explainability} \\
            
            % 中间横线 (分组画线)
            \cline{2-5} \cline{6-9} \cline{10-12}
            \noalign{\vspace{1pt}}
            
            % 第2行子标题
             & Acc & F1 & Real & Fake & Hum & Obj & Sce & Ani & Rel & Exp & Com \\
            
            \hline % 表头下方的黑线
            \noalign{\vspace{3pt}}
        
            \quad Effort \textit{(ICML 2025)}~\cite{yan2024effort} & 56.7 & 41.9 & 31.3 & 82.1 & 63.0 & 53.3 & 53.0 & 57.4 & - & - & - \\
            \quad NPR \textit{(CVPR 2024)}~\cite{tan2024rethinking} & 60.4 & 60.9 & 61.7 & 59.1 & 43.7 & 75.5 & 52.6 & 69.7 & - & - & - \\
            \quad AIDE \textit{(ICLR 2025)}~\cite{yan2024sanity} & 73.9 & 78.4 & 94.5 & 53.4 & 56.6 & \underline{84.0} & 68.7 & 86.3 & - & - & - \\
            \hline
            \noalign{\vspace{3pt}}
            \quad GPT-4o~\cite{gpt4o} & 69.5 & 74.3 & 88.4 & 50.5 & 66.5 & 67.8 & 74.4 & 69.0 & 92.8 & 81.4 & 50.8 \\
            \quad GPT-5~\cite{openai_gpt5_system_card_2025} & 73.7 & 77.9 & 92.7 & 54.7 & 65.6 & 80.3 & 73.5 & 75.4 & 93.1 & 84.3 & 67.2 \\
            \quad Gemini-2.5~\cite{google2025gemini} & 73.3 & \underline{78.6} & \underline{98.4} & 48.2 & 73.7 & 75.7 & 72.6 & 71.1 & \underline{93.6} & \underline{90.4} & \underline{76.5} \\
            \hline
            \noalign{\vspace{3pt}}
            
            \quad DeepSeekVL2-7B~\cite{wu2024deepseekvl2mixtureofexpertsvisionlanguagemodels} & 50.2 & 66.8 & \textbf{99.7} & 0.4 & 50.1 & 50.2 & 50.1 & 50.0 & 77.9 & 72.3 & 33.5 \\
            \quad InternVL3-8B~\cite{zhu2025internvl3exploringadvancedtraining} & 52.0 & 60.4 & 73.3 & 30.7 & 49.8 & 54.1 & 51.6 & 52.4 & 69.4 & 62.6 & 36.5 \\
            \quad InternVL3-38B~\cite{zhu2025internvl3exploringadvancedtraining} & 56.0 & 66.7 & 88.3 & 23.7 & 54.5 & 59.4 & 53.5 & 56.5 & 70.4 & 63.1 & 35.7 \\
            \quad Qwen2.5-VL-7B~\cite{bai2025qwen25vltechnicalreport} & 61.4 & 71.3 & 96.2 & 26.6 & 71.4 & 61.0 & 59.1 & 54.0 & 91.3 & 84.3 & 44.4 \\
            \quad Qwen2.5-VL-32B~\cite{bai2025qwen25vltechnicalreport} & 67.3 & 74.7 & 96.4 & 38.3 & 75.9 & 67.6 & 64.8 & 61.0 & 92.2 & 84.4 & 44.5 \\
            \hline
            \noalign{\vspace{3pt}}
            \quad Skyra \textit{(CVPR 2026)}~\cite{li2025skyra} & 68.9 & 73.6 & 86.7 & 51.0 & 68.4 & 72.8 & 69.3 & 64.9 & 84.9 & 79.5 & 53.5 \\
            \quad SIDA \textit{(CVPR 2025)}~\cite{huang2025sida} & 65.3 & 68.7 & 76.3 & 54.3 & 69.3 & 59.5 & 68.3 & 64.0 & 73.1 & 61.6 & 40.0 \\
            \quad BusterX++~\cite{wen2025busterx++} & 76.2 & 75.3 & 72.7 & 79.7 & 78.0 & 72.5 & \underline{79.9} & 74.4 & 74.1 & 69.0 & 46.5 \\
            \quad FakeVLM \textit{(NIPS 2025)}~\cite{wen2025spot} & \underline{77.0} & 70.4 & 54.8 & \textbf{99.3} & \underline{78.3} & 72.0 & 69.4 & \underline{88.3} & 31.0 & 20.4 & 13.1 \\
            
            % --- 高亮最后一行 (Faker1-R1) ---
            \quad \textbf{FakeVLM-R1} & \textbf{95.5} & \textbf{95.5} & 95.9 & \underline{95.2} & \textbf{95.1} & \textbf{95.4} & \textbf{93.6} & \textbf{98.0} & \textbf{95.1} & \textbf{91.9} & \textbf{77.7} \\
            
            \hline % 底部黑线
        \end{tabular}
    }
    \vspace{-5pt}
\end{table*}

\subsection{Explainable Synthetic Detection}
This section comprehensively evaluates FakeVLM-R1. We first utilize FakeClue++ to assess its detection performance and logical depth. Next, LOKI experiments test its resilience against shortcut overfitting, followed by a fine-grained visualization investigating improvement in class biases and explanatory hallucinations.

\textit{1) Detection Task on FakeClue++:} The fundamental capability of a synthetic detection model lies in the effective forensic discrimination between forged and authentic images. As shown in Table \ref{tab:fakeclue++_eval}, FakeVLM-R1 demonstrates exceptional performance on FakeClue++, achieving an accuracy of 95.5\% and an F1 score of 95.5\%. This not only surpasses binary classification models and general-purpose LMMs, but also exhibits a paradigm-level advantage when compared in-depth with existing interpretable synthetic detection models, such as Skyra, SIDA, BusterX++, and the predecessor FakeVLM. Table \ref{tab:fakeclue++_eval} reveals that existing interpretable models fall into a dilemma of real/fake discrimination imbalance. For instance, the accuracies of Skyra and SIDA on Fake images hover merely around 50\%, indicating a near-complete loss of sensitivity to forgery features. Conversely, while the predecessor FakeVLM can precisely localize artifacts, with fake accuracy reaching 99.3\%, its lack of physical priors results in a severe over-rejection bias on authentic images, with real accuracy of only 54.8\%, making it highly susceptible to mechanically misinterpreting natural noise as generative traces. In contrast, FakeVLM-R1 achieves consistent gains across benchmarks. By introducing a bidirectional dialectical reasoning mechanism and deeply integrating physical world knowledge, the model is compelled to execute a self-adversarial process of "authenticity counter-proof" when confronted with visual clues. This mechanism endows the model with profound discriminative confidence, enabling it to achieve a near-perfect equilibrium and outstanding performance between real-world images (95.9\%) and synthetic images (95.2\%). It also achieves high detection accuracy across the four core categories (Human, Object, Scene, and Animal). Furthermore, this demonstrates that although FakeClue++ comprises a smaller volume of data, a synthetic detection reasoning model trained based on critical thinking can effectively achieve ideal results with significantly less data.

\textit{2) Reasoning Task on FakeClue++:} In the evaluation of the reasoning as shown in Table \ref{tab:fakeclue++_eval}, FakeVLM-R1 achieves outstanding scores of 95.1, 91.9, and 77.7 across three dimensions: Relevance, Logicality, and Completeness, respectively. This fully corroborates the profound facilitating effect of high-quality reasoning on detection accuracy. Revisiting FakeVLM, it was constrained by a mechanical probabilistic mapping characterized by a "conclusion-first, explanation-subsequent" paradigm, leaving it deeply mired in severe explanatory hallucinations (with a Rel of merely 31.0 and a Com of 13.1), and resulted in its high false positive rate on authentic images. In contrast, by integrating GRPO with an explicit CoT, FakeVLM-R1 executes a counter-proof prior to generating a final verdict, which not only precipitates a qualitative leap in the logical depth and granular coverage of the explanations, but also fundamentally underpins the remarkable detection accuracy through a rigorous, forensic-level reasoning process. Furthermore, an analysis of comparable open-source interpretable baselines, such as Skyra and BusterX++, reveals a pervasive inability to produce exhaustive reasoning trajectories, as evidenced by Completeness scores consistently falling below 55, and there is a discernible correlation between explanatory capability and detection performance. FakeVLM-R1 overwhelmingly surpasses top-tier closed-source LMMs equipped with massive prior knowledge, such as GPT-5 and Gemini-2.5, in explanations. Ultimately, these results compellingly demonstrate that anchoring synthetic image detection in critical thinking actualizes a genuine co-evolution of logical explication and precise discrimination.

\textit{3) Results on LOKI:} To further validate the effectiveness of our model, we conducted evaluation on LOKI. As shown in Table \ref{tab:loki_eval}, in the Real/Fake discrimination task, FakeVLM-R1 achieves an accuracy of 88.25\%, securing the top position. Crucially, the LOKI dataset introduces a Yes/No question-answering paradigm that necessitates dynamic logical conversion. This serves as a touchstone for determining whether the model possesses genuine semantic understanding or is merely relying on rote memorization. Table \ref{tab:loki_eval} reveals that FakeVLM lacks a deep reasoning process and suffers a catastrophic performance collapse on this task. Its accuracy plummets from 85.76\% in direct discrimination to 66.75\% in the Yes/No paradigm, exposing that its historically superior performance relied heavily on shortcut overfitting to fixed output formats. In contrast, FakeVLM-R1 exhibits profound logical resilience, maintaining an exceptionally high accuracy of 84.64\% on the Yes/No task. It not only overcomes the critical flaw of format overfitting but also surpasses GPT-5 (80.92\%) with massive generalized knowledge. This demonstrates that, by virtue of the bidirectional dialectical reasoning mechanism, the model is capable of independently executing the foundational logical deductions of both a "forgery hypothesis" and an "authenticity counter-proof" within its chain of thought prior to generating a final answer. By first accurately anchoring the intrinsic authenticity of the image and subsequently adjusting its output dynamically based on the specific instruction, the model actualizes a genuine paradigm shift from rote perceptual matching to robust causal reasoning.

\begin{table}[t]
    \centering
    \renewcommand{\arraystretch}{1.2}
    \caption{Detection accuracy and explainability metrics evaluated on LOKI. All results are expressed as percentages (\%). \textbf{Bold} and \underline{underlined} values indicate the best and second-best results, respectively.}
    \label{tab:loki_eval}
    \vspace{-5pt}
    \resizebox{\linewidth}{!}{
        \begin{tabular}{lccccccc}
            \hline
            % --- 复合表头 ---
            \multicolumn{1}{c}{Method} & \multicolumn{2}{c}{Real / Fake Accuracy} & \multicolumn{2}{c}{Yes / No Accuracy} & \multicolumn{3}{c}{Explainability} \\
            \cline{2-3} \cline{4-5} \cline{6-8}
            \noalign{\vspace{1pt}}
            % 第2行子标题
             & Acc & F1 & Acc & F1 & Rel & Exp & Com \\
            \hline
            \noalign{\vspace{1pt}}
            Effort~\cite{yan2024effort} & 75.28 & 56.71 & - & - & - & - & - \\
            NPR~\cite{tan2024rethinking} & 77.98 & 67.52 & - & - & - & - & - \\
            AIDE~\cite{yan2024sanity} & 79.72 & 76.86 & - & - & - & - & - \\
            \hline
            \noalign{\vspace{1pt}}
            gpt-4o~\cite{gpt4o} & 76.39 & 73.87 & 77.96 & 75.26 & 92.21 & 90.50 & 39.37 \\
            gpt-5~\cite{openai_gpt5_system_card_2025} & 76.83 & 75.60 & \underline{80.92} & \underline{83.45} & \underline{97.24} & 93.26 & \underline{79.95} \\
            gemini-2.5~\cite{google2025gemini} & 68.58 & 69.89 & 76.00 & 72.27 & 96.95 & \underline{94.40} & 78.27 \\
            \hline
            \noalign{\vspace{1pt}}
            DeepDeekVL2-7B~\cite{wu2024deepseekvl2mixtureofexpertsvisionlanguagemodels} & 36.64 & 53.63 & 58.16 & 53.43 & 74.98 & 63.65 & 33.74 \\
            InternVL3-8B~\cite{zhu2025internvl3exploringadvancedtraining} & 53.26 & 55.58 & 57.44 & 63.21 & 79.04 & 73.41 & 45.32 \\
            InternVL3-38B~\cite{zhu2025internvl3exploringadvancedtraining} & 51.52 & 57.58 & 58.49 & 60.48 & 85.34 & 71.88 & 48.10 \\
            Qwen2.5-VL-7b~\cite{bai2025qwen25vltechnicalreport} & 41.43 & 53.30 & 44.79 & 31.69 & 72.96 & 58.89 & 46.32 \\
            Qwen2.5-VL-32b~\cite{bai2025qwen25vltechnicalreport} & 55.27 & 61.41 & 46.88 & 28.00 & 90.27 & 68.56 & 68.43 \\
            \hline
            \noalign{\vspace{1pt}}
            Skyra~\cite{li2025skyra} & 46.91 & 56.12 & 60.05 & 60.40 & 87.07 & 75.69 & 49.11 \\
            SIDA~\cite{huang2025sida} & 59.01 & 61.68 & - & - & 83.89 & 77.86 & 44.46 \\
            BusterX++~\cite{wen2025busterx++} & 79.03 & 67.03 & 66.54 & 54.30 & 90.18 & 72.97 & 55.18 \\
            FakeVLM~\cite{wen2025spot} & \underline{85.76} & \underline{76.30} & 66.75 & 55.05 & 45.44 & 29.62 & 20.33 \\
            \textbf{FakeVLM-R1} & \textbf{88.25} & \textbf{82.21} & \textbf{84.64} & \textbf{83.98} & \textbf{98.53} & \textbf{94.80} & \textbf{80.83} \\
            \hline
        \end{tabular}
    }
\end{table}

\begin{figure}[!t]
\centering
\includegraphics[width=1.0\columnwidth]{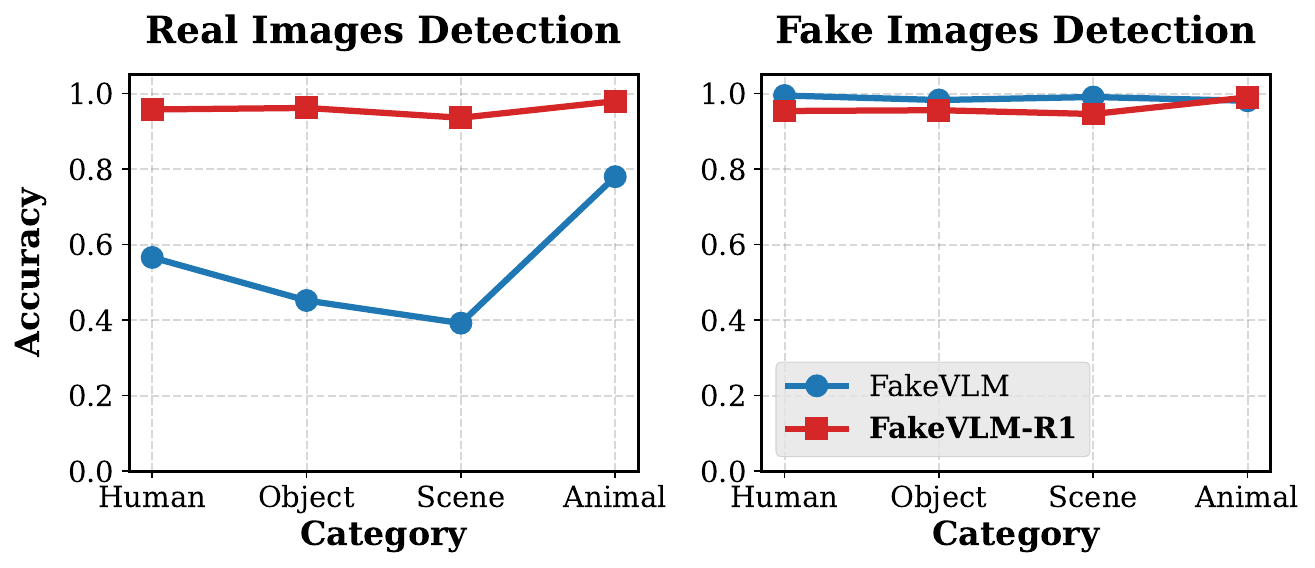}
\caption{Fine-grained comparison in FakeClue++. Performance comparison between FakeVLM-R1 and FakeVLM across diverse image categories and ground-truth labels.}
\label{line_comparison}
\end{figure}

\begin{figure*}[!t]
\centering
\includegraphics[width=2.0\columnwidth]{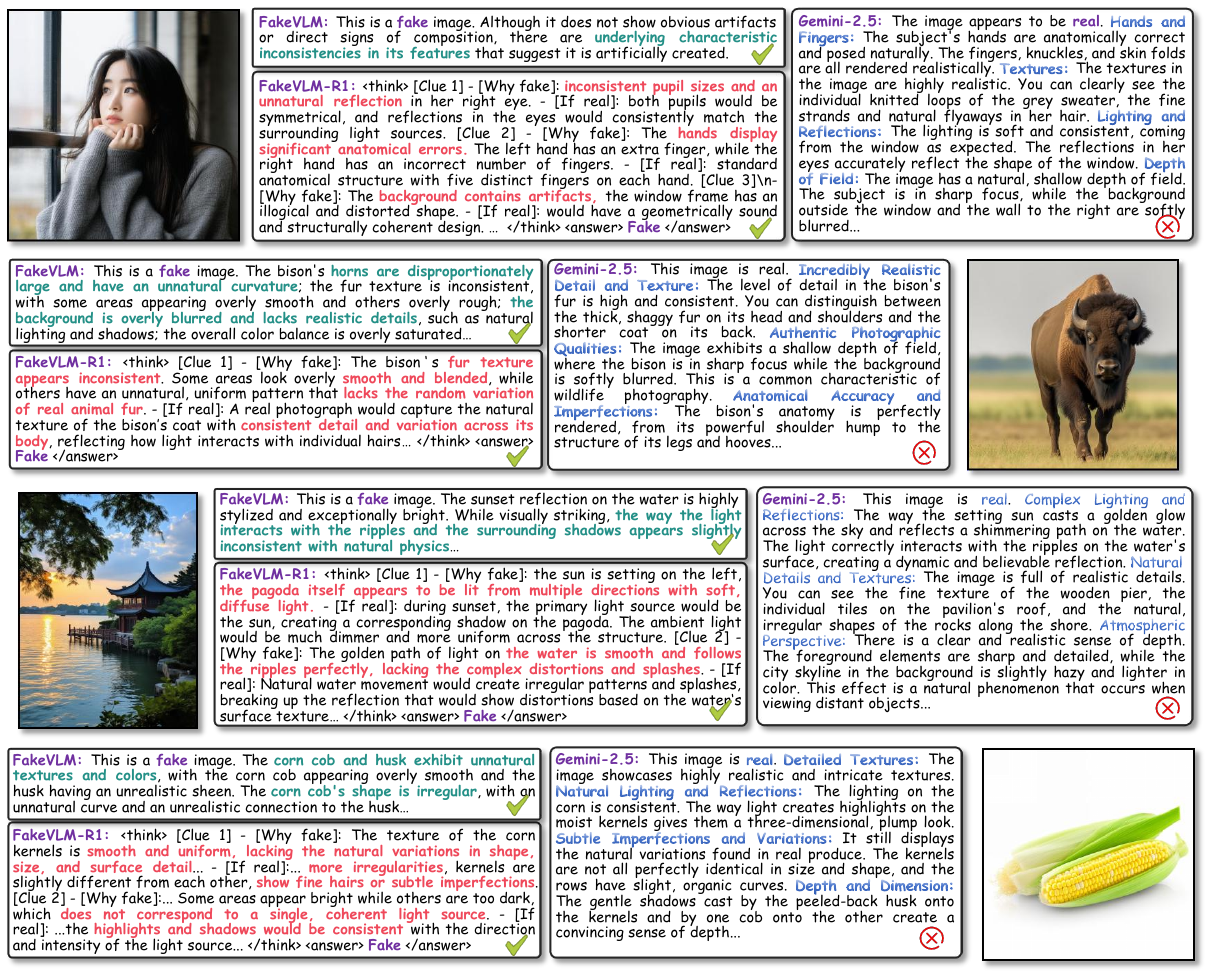}
\caption{Comparison of reasoning processes on fake Samples among FakeVLM-R1, FakeVLM and Gemini-2.5.}
\label{fake_comparison}
\end{figure*}

\begin{figure*}[!t]
\centering
\includegraphics[width=2.0\columnwidth]{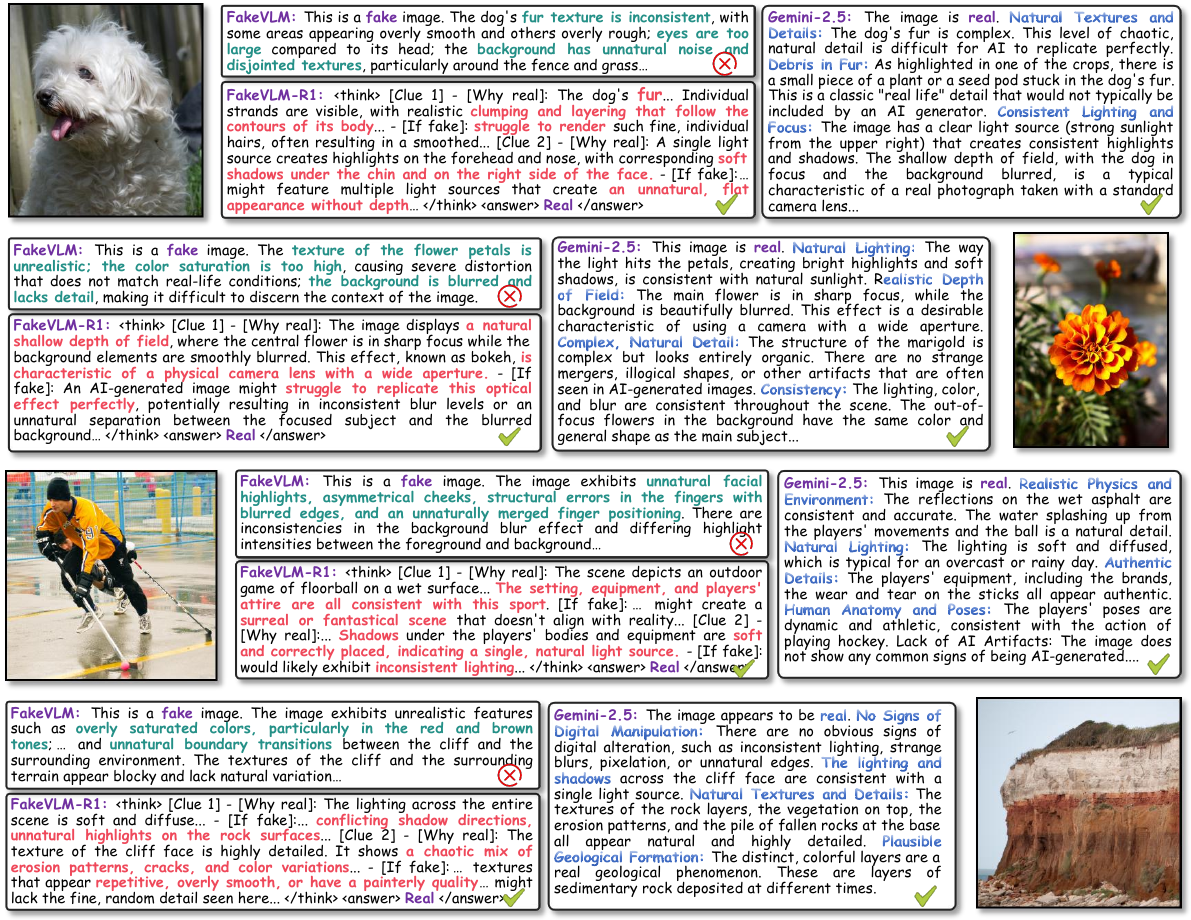}
\caption{Comparison of reasoning processes on real Samples among FakeVLM-R1, FakeVLM and Gemini-2.5.}
\label{real_comparison}
\end{figure*}

\textit{4) Fine-grained analysis:} To provide a more intuitive demonstration of the discriminative differences and reasoning depth across four categories, we conduct an analysis integrating quantitative and qualitative cases. During the fine-grained category evaluation in Fig.~\ref{line_comparison}, FakeVLM exhibits an extreme imbalance in real/fake discrimination: although it performs adequately in fake image detection, it suffers a catastrophic performance degradation in real image detection, with its accuracy in the Scene category plummeting to approximately 0.4. In contrast, FakeVLM-R1 stably maintains a high accuracy approaching 1.0 across all four categories, resolving the class bias and high false positive rate. This performance chasm is fully explicated within the qualitative cases as shown in Fig.~\ref{fake_comparison} and Fig.~\ref{real_comparison}. When confronted with synthetic images, the explanations generated by FakeVLM frequently stall at superficial and ambiguous descriptions, such as citing textures that are overly smooth or rough. FakeVLM-R1, however, demonstrates forensic-level bidirectional dialectical thinking; it not only precisely localizes multi-dimensional artifacts, such as inconsistent pupil reflections and structural finger anomalies, but also compellingly invokes standard anatomical structures and optical physics priors to conduct counterfactual deductions. When evaluating authentic images, FakeVLM falls into severe explanatory hallucinations, mechanically fabricating forgery evidence for genuine animal fur or flower close-ups, which ultimately culminates in complete misjudgments. Conversely, FakeVLM-R1 accurately captures and comprehends real-world physical attributes, such as optical shallow depth of field, coherent single light source distributions, and complex natural weathering textures. By comparatively deducing the common defects of AI generative models when rendering these physical details, it ultimately yields logically self-consistent and highly confident correct verdicts.

\subsection{Generalization and Robustness}
In this section, we validate the generalization and robustness of the proposed method. We conduct experiments on MMFakeBench and DMimage, and utilize the image data from these benchmarks to demonstrate the generalization capability of FakeVLM-R1. Moreover,  to evaluate the robustness, we subject the images from the FakeClue++ dataset to a diverse set of format conversions, and compare it with FakeVLM.

\textit{1) Results on MMFakeBench:} MMFakeBench is a multimodal dataset of 11,000 image-text pairs spanning textual, visual, and cross-modal distortions. Generated via models like ChatGPT and Midjourney, it includes fine-grained forgeries alongside authentic control data to evaluate model robustness in complex, mixed-source scenarios. Experimental analysis on MMFakeBench as shown in Table~\ref{MMFakeBench}, reveals that traditional binary classifiers (e.g., NPR, AIDE), despite excelling in identifying real images, suffer from severe bias on forged samples. While these data-driven approaches learn artifact features, they often lack a concrete understanding of image semantics, resulting in poor generalization. Conversely, interpretable methods (e.g., BusterX++, FakeVLM) effectively detect forgeries but sacrifice accuracy on authentic images, leading to a high false positive rate. In contrast, FakeVLM-R1 achieves the optimal performance trade-off: it maintains a 98.2\% accuracy on real images while significantly elevating the forgery detection rate to 93.2\%, culminating in an overall accuracy of 96.3\% and demonstrating superior generalization capability.

\begin{table}[t]
    \centering
    \renewcommand{\arraystretch}{1.0}
    \caption{Detection accuracy metrics evaluated on MMFakeBench. All results are expressed as percentages (\%). \textbf{Bold} and \underline{underlined} values indicate the best and second-best results, respectively.}
    \label{MMFakeBench}
    \vspace{-5pt}
    \resizebox{\linewidth}{!}{
        \begin{tabular}{lcccccc}
            \hline % 顶部黑线
            
            % --- 复合表头 ---
            \multicolumn{1}{c}{{Method}} & \multicolumn{2}{c}{Real} & \multicolumn{2}{c}{Fake} & \multicolumn{2}{c}{Overall} \\
            
            % 中间横线
            \cline{2-7}
            
            % 第2行子标题
             & Acc & F1 & Acc & F1 & Acc & F1 \\
            
            \hline % 表头下方的黑线
            
            \quad Effort~\cite{yan2024effort} & 94.7 & 92.6 & 83.7 & 86.9 & 90.6 & 89.8 \\
            \quad NPR~\cite{tan2024rethinking} & \underline{97.7} & 88.5 & 57.4 & 72.6 & 83.8 & 80.6 \\
            \quad AIDE~\cite{yan2024sanity} & 97.3 & 85.9 & 46.6 & 63.1 & 79.5 & 74.5 \\
            % \quad DeepSeekVL-7B & \textbf{100.0} & 77.1 & 1.2 & 2.4 & 62.9 & 39.8 \\
            % \quad InternVL3-8B & 71.0 & 67.3 & 33.3 & 36.7 & 56.8 & 52.0 \\
            % \quad InternVL3-38B & 84.6 & 74.8 & 30.7 & 39.2 & 64.4 & 57.0 \\
            % \quad Qwen2.5-VL-7B & 89.4 & 81.4 & 48.9 & 58.6 & 74.3 & 70.0 \\
            % \quad Qwen2.5-VL-32B & 95.1 & 89.0 & 69.1 & 77.9 & 85.3 & 83.5 \\
            \quad Skyra~\cite{li2025skyra} & 71.1 & 79.2 & 86.1 & 73.5 & 76.7 & 76.4 \\
            \quad SIDA~\cite{huang2025sida} & 90.6 & 88.8 & 77.5 & 80.3 & 85.7 & 84.5 \\
            \quad BusterX++~\cite{wen2025busterx++} & 86.1 & 91.9 & \underline{93.0} & 88.6 & 90.6 & 90.3 \\
            \quad FakeVLM~\cite{wen2025spot} & 93.6 & \underline{93.6} & 89.2 & \underline{89.3} & \underline{92.0} & \underline{91.4} \\
            \quad \textbf{FakeVLM-R1} & \textbf{98.2} & \textbf{97.1} & \textbf{93.2} & \textbf{95.0} & \textbf{96.3} & \textbf{96.0} \\
            
            \hline % 底部黑线
        \end{tabular}
    }
    \vspace{-5pt}
\end{table}

\textit{2) Results on DMimage:} DMimage is a large-scale binary classification benchmark for synthetic image detection. It features diverse synthetic images from GAN, diffusion, and Transformer models (1,000 per model), complemented by a control group of 5,000 real images from COCO, ImageNet, and UCID. Experiments on the DMimage dataset as shown in Table~\ref{DMimage}, demonstrate that FakeVLM-R1 consistently outperforms the predecessor model, and other expert models across all metrics. Notably, in the challenging task of fake image detection, FakeVLM-R1 achieves a significant performance leap, raising the accuracy from 89.7\% to 94.6\% and establishing a new state-of-the-art Overall Accuracy of 96.5\%. These results strongly evidence that FakeVLM-R1 not only inherits the multimodal interpretability of its predecessor but also leverages the 'bidirectional dialectic' strategy to more sensitively capture subtle artifacts in unseen generative domains.

\begin{table}[t]
    \centering
    \renewcommand{\arraystretch}{1.0}
    \caption{Detection accuracy metrics evaluated on DMimage. All results are expressed as percentages (\%). \textbf{Bold} values indicate the best results.}
    \label{DMimage}
    \vspace{-5pt}
    \resizebox{\linewidth}{!}{
        % 修改点1：去除了列定义中的 "|"，变为纯 lcccccc
        \begin{tabular}{lcccccc}
            \hline % 顶部黑线
            % --- 复合表头 ---
            % 修改点2：为了让 "Method" 在这两行既垂直居中又水平居中，
            % 我们用 \multicolumn{1}{c}{...} 将其包裹，强制该单元格水平居中
            \multicolumn{1}{c}{{Method}} & \multicolumn{2}{c}{Real} & \multicolumn{2}{c}{Fake} & \multicolumn{2}{c}{Overall} \\
            % 中间横线
            \cline{2-7}
            % 第2行子标题
             & Acc & F1 & Acc & F1 & Acc & F1 \\
            \hline % 表头下方的黑线
            \quad CNNSpot~\cite{wang2020cnn} & 87.8 & 88.4 & 28.4 & 44.2 & 40.6 & 43.3 \\
            \quad Gram-Net~\cite{liu2020global} & 62.8 & 54.1 & 78.8 & 88.1 & 67.4 & 79.4 \\
            \quad Fusing~\cite{ju2022fusing} & 87.7 & 86.1 & 15.5 & 27.2 & 40.4 & 36.5 \\
            \quad LNP~\cite{bi2023detecting} & 63.1 & 67.4 & 56.9 & 72.5 & 58.2 & 68.3 \\
            \quad UnivFD~\cite{ojha2023towards} & 89.4 & 88.3 & 44.9 & 61.2 & 53.9 & 60.7 \\
            \quad AntifakePrompt ~\cite{chang2023antifakeprompt} & 91.3 & 92.5 & 89.3 & 91.2 & 90.6 & 91.2 \\
            \quad SIDA~\cite{huang2025sida} & 92.9 & 93.1 & 90.7 & 91.0 & 91.8 & 92.4 \\
            \quad FakeVLM~\cite{wen2025spot} & 98.2 & 99.1 & 89.7 & 94.6 & 94.0 & 94.3 \\
            \quad \textbf{FakeVLM-R1} & \textbf{98.4} & \textbf{99.2} & \textbf{94.6} & \textbf{96.4} & \textbf{96.5} & \textbf{96.4} \\
            
            \hline % 底部黑线
        \end{tabular}
    }
    \vspace{-5pt}
\end{table}

\textit{3) Robustness Test:} To evaluate the model's robustness in complex real-world scenarios, we applied various perturbations to the input images (including JPEG compression, resizing, Gaussian noise, geometric transformations, contrast adjustment, and blurring). As shown in Table~\ref{tab:robustness}, the reported values represent the percentage change in performance, with 'Overall' denoting the mean of the absolute changes. Experimental results demonstrate that FakeVLM-R1 achieves a substantial leap in its resilience against perturbations. In terms of overall stability, the average fluctuation in the total accuracy of FakeVLM-R1 is a mere 3.63\%, significantly outperforming the 9.44\% fluctuation observed in the predecessor. Crucially, FakeVLM-R1 completely resolves the critical vulnerability of traditional models, which are prone to "catastrophic collapse" on real images when subjected to perturbations. An analysis of the data reveals that the original FakeVLM, which relies heavily on superficial pixel-level matching, suffers a precipitous drop in Real accuracy by -94.70\% and -28.68\% when subjected to blurring and contrast reduction , respectively. This leads to a severe overall fluctuation of 15.85\% for this category. In stark contrast, FakeVLM-R1 exhibits exceptional resilience, drastically compressing the overall average fluctuation for the Real category to a mere 2.90\%. This compellingly demonstrates that, benefiting from the internalization of physical commonsense and the bidirectional dialectical reasoning mechanism, the model remains capable of invoking deep logical principles to construct an "authenticity counter-proof" even when low-level texture details are compromised. Furthermore, under highly destructive JPEG 70 compression, FakeVLM-R1 successfully halves the accuracy degradation, reducing the loss from -20.07\% to -10.29\%. In conclusion, elevating synthetic image detection from passive feature perception to active causal reasoning not only raises the upper bound of detection performance but also endows the model with highly practical robustness against unseen perturbations.
\begin{table}[t]
    \centering
    \renewcommand{\arraystretch}{1.0}
    \caption{Robustness evaluation under different perturbations. All results are expressed as percentages (\%).}
    \label{tab:robustness}
    \vspace{-5pt}
    \resizebox{\linewidth}{!}{
        % 定义7列：1列左对齐(Method)，6列居中(数据)
        \begin{tabular}{lcccccc}
            \hline % 顶部黑线
            % --- 复合表头 ---
            % 第一行：Method 占1列，两个模型各占3列
            \multicolumn{1}{c}{Conversion} & \multicolumn{3}{c}{FakeVLM-R1} & \multicolumn{3}{c}{FakeVLM} \\
            % 中间横线，只覆盖数据列(2-7)
            \cline{2-7}
            % 第二行：具体的指标
             & Acc & Real & Fake & Acc & Real & Fake \\
            \hline % 表头下方的黑线
            
            \quad JPEG 70 & -10.29 & 0.63 & -21.28 & -20.07 & -4.66 & -29.56 \\
            \quad JPEG 80 & -6.57 & -0.05 & -13.14 & -19.68 & -3.56 & -29.56 \\
            \quad Resize 0.5 & -8.41 & -7.87 & -8.93 & -9.62 & -14.57 & -2.00 \\
            \quad Resize 0.75 & -3.87 & -4.38 & -3.36 & -4.13 & -8.37 & -1.10 \\
            \quad Gaussian 10 & -0.61 & 1.67 & -2.89 & -4.35 & -10.05 & -1.21 \\
            \quad Gaussian 5 & -0.11 & 0.94 & -1.16 & -3.25 & -6.67 & -1.36 \\
            \quad Flip horizontal & -0.17 & -0.21 & -0.11 & 0.78 & 1.92 & 0.15 \\
            \quad Rotate 15 & -3.25 & -2.97 & -3.52 & -3.43 & -1.20 & -2.22 \\
            \quad Sharpen 1.5 & -1.03 & 1.20 & -3.26 & 2.64 & 8.31 & -0.50 \\
            \quad Contrast 0.7 & -0.03 & -0.16 & 0.11 & -9.87 & -28.68 & 0.50 \\
            \quad Contrast 1.3 & -2.62 & 2.03 & -7.30 & 2.30 & 7.58 & -0.61 \\
            \quad Blur 3 & -6.61 & -12.67 & -0.47 & -33.18 & -94.70 & 0.76 \\
            \hline
            \quad Overall & 3.63 & 2.90 & 5.46 & 9.44 & 15.85 & 5.79 \\
            
            \hline % 底部黑线
        \end{tabular}
    }
    \vspace{-5pt}
\end{table}

\subsection{Ablation Experiments}
To rigorously evaluate the individual contributions and synergistic effects of the core components within FakeVLM-R1—namely, SFT, the Critical Thinking (CT) strategy, GRPO, and the GRPO reward functions—we conducted a systematic ablation study utilizing Qwen2.5-VL as the base model. Except for the specific components being ablated, all other experimental setups, including training configurations and datasets, are kept strictly identical.

\textit{1) Training Strategy Analysis:} Experimental results in Table \ref{tab:ablation_study1} reveals that the unaligned base model suffers from severe class bias. By incorporating SFT, the model acquires fundamental visual forensic priors. Consequently, the overall accuracy surges to 86.3\% and the explanation completeness improves to a score of 75.5. This underscores the critical role of the supervised fine-tuning phase in establishing foundational perceptual capabilities during the cold start. Conversely, directly applying reinforcement learning without prior supervised fine-tuning manages to elevate the overall accuracy to 92.7\% through exploration. However, due to the absence of a foundational knowledge base, the completeness of its generated explanations is notably constrained, dropping to a score of 63.5. A more pivotal finding of this study centers on the decisive impact of the critical thinking strategy. When this strategy is excluded from the complete supervised fine-tuning and reinforcement learning pipeline, relying solely on conventional unidirectional reasoning effectively captures artifacts, as evidenced by a 94.3\% accuracy on fake images. Nevertheless, this approach triggers severe side effects characterized by overfitting and false positives, causing the accuracy on authentic images to plummet to 89.0\%. Such degradation exposes a lack of profound discriminative confidence in complex scenarios. In stark contrast, fully integrating the explicit critical thinking strategy compels the model to engage in a self-adversarial process of proposing a forgery hypothesis and constructing an authenticity counter-proof during the reasoning phase. This mechanism perfectly rectifies the imbalance in real-versus-fake discrimination, restoring the accuracy on real images to 95.9\% and propelling the overall accuracy to an optimal 95.5\%. Furthermore, it achieves remarkable score across the respective metrics of relevance, explainability, and completeness. This compellingly corroborates that the critical thinking strategy acts not merely as a catalyst for elevating the upper bound of detection performance, but rather as the foundational cornerstone ensuring the model's logical self-consistency and forensic-level depth of analysis.

\begin{table}[t]
    \centering
    \renewcommand{\arraystretch}{1.2}
    \caption{Ablation study on different training strategies. All results are expressed as percentages (\%). \textbf{Bold} and \underline{underlined} values indicate the best and second-best results, respectively.}
    \label{tab:ablation_study1}
    \vspace{-5pt}
    \resizebox{\linewidth}{!}{
        \begin{tabular}{lcccccc}
            \hline
            % --- 复合表头 ---
            \multicolumn{1}{c}{Method} & \multicolumn{3}{c}{Total Detection} & \multicolumn{3}{c}{Explainability} \\
            \cline{2-4} \cline{5-7}
            % 第2行子标题
             & Real & Fake & Total & Rel & Exp & Com \\
            \hline
            -Base Model & \textbf{96.2} & 26.6 & 61.4 & 91.3 & 84.3 & 44.4 \\
            +SFT & 84.5 & 88.0 & 86.3 & 93.5 & 90.5 & 75.5 \\
            +GRPO \& CT & 95.2 & 88.2 & \underline{92.7} & 90.3 & 92.1 & 63.5 \\
            +SFT \& GRPO & 89.0 & \underline{94.3} & 91.6 & \underline{96.3} & \underline{93.1} & \underline{78.4} \\
            +SFT \& GRPO \& CT & \underline{95.9} & \textbf{95.2} & \textbf{95.5} & \textbf{98.5} & \textbf{94.1} & \textbf{80.8} \\
            \hline
        \end{tabular}
    }
\end{table}

\textit{2) GRPO Reward Analysis:}To investigate the impact of the composite reward system, we ablated the reward components within the GRPO as shown in Table \ref{tab:ablation_study2}. As shown in the table, retaining only accuracy and format rule rewards ($R_{acc} \& R_{fmt}$) yields a 92.4\% overall accuracy but causes a catastrophic collapse in explainability metrics, indicating that simple binary feedback degrades the model into a probabilistic fitting machine devoid of reasoning. Introducing structural and logical rewards effectively resolves this: removing the dialectical structure reward (remove $R_{struc}$) degrades forgery detection and explanation relevance (Rel), confirming the necessity of paired forgery hypothesis and authenticity counter-proof outputs to anchor visual clues. Similarly, removing the logical consistency reward (remove $R_{logic}$) causes a sharp 10.1-point drop in logicality (Exp), highlighting the semantic judge's core role in preventing logical gaps in reasoning. Furthermore, removing the cosine length reward (remove $R_{len}$) triggers a comprehensive decline across all metrics, demonstrating that dynamic length regulation effectively forces the model to allocate more computation budget for deep exploration on complex samples. Ultimately, by aggregating all these constraints, the model achieves an optimal co-evolution of detection accuracy and explanation completeness.

\begin{table}[t]
    \centering
    \renewcommand{\arraystretch}{1.2}
    \caption{Ablation study on different GRPO reward function designs. All results are expressed as percentages (\%). \textbf{Bold} and \underline{underlined} values indicate the best and second-best results, respectively.}
    \label{tab:ablation_study2}
    \vspace{-5pt}
    \resizebox{\linewidth}{!}{
        \begin{tabular}{lcccccc}
            \hline
            % --- 复合表头 ---
            \multicolumn{1}{c}{Method} & \multicolumn{3}{c}{Total Detection} & \multicolumn{3}{c}{Explainability} \\
            \cline{2-4} \cline{5-7}
            % 第2行子标题
             & Real & Fake & Total & Rel & Exp & Com \\
            \hline
            \noalign{\vspace{3pt}}
            $R_{acc} \ \& \ R_{fmt}$ & 94.1 & 90.8 & 92.4 & 16.3 & 10.7 & 8.0 \\
            remove $R_{len}$ & 95.7 & 93.4 & 94.5 & 96.5 & 81.2 & 74.9 \\
            remove $R_{struc}$ & \textbf{96.3} & 92.2 & 94.3 & 91.5 & \underline{92.8} & \underline{78.7} \\
            remove $R_{logic}$ & \underline{96.2} & \underline{94.3} & \underline{95.3} & \underline{97.0} & 84.0 & 78.2 \\
            Ours & 95.9 & \textbf{95.2} & \textbf{95.5} & \textbf{98.5} & \textbf{94.1} & \textbf{80.8} \\
            \hline
        \end{tabular}
    }
\end{table}

\section{Conclusions and future work}
This paper aims to propel synthetic image detection from passive visual perception to active forensic-level causal reasoning by proposing an advanced detection framework, FakeVLM-R1, alongside a complementary high-quality interpretable benchmark, FakeClue++. At the data construction level, we posit that describing the inherent physical laws of authentic images is significantly more intuitive and efficient than exhaustively enumerating ever-changing forgery traces. Consequently, FakeClue++ innovatively incorporates physical prior annotations for real images, providing the model with unified and robust "authenticity anchors" at minimal annotation costs. At the methodology level, to address the critical pain points of existing SFT-based interpretable models—namely, explanatory hallucinations and high false positive rates on real images, we abandon traditional unidirectional mapping and innovatively integrate GRPO with the CoT strategy. By compelling the model to execute a bidirectional dialectical game of "forgery/authenticity hypothesis clues" and "authenticity/forgery counter-proof" during the reasoning phase, constrained by a multi-dimensional "structure-semantics-length" reward system, FakeVLM-R1 encourages the model to leverage physical priors, such as lighting consistency, anatomical perspective. It successfully internalizes this physical commonsense into deep discriminative confidence. Extensive experiments across multiple mainstream benchmarks, including FakeClue++ and LOKI, demonstrate that FakeVLM-R1 not only achieves high overall detection accuracy (e.g., 95.5\% on FakeClue++) and substantially reduces class bias, but also yields improvements over prior methods in cross-domain generalization, robustness against perturbations, and the logical self-consistency of explanations.

Although FakeVLM-R1 has achieved remarkable progress in static image forensics, the rapid evolution of generative AI technologies presents vast opportunities for future exploration. Specifically, our future work will primarily unfold across three dimensions. First, extending to temporal and multimodal domains. We plan to leverage inter-frame consistency and audio-visual synchronization logic as higher-order physical world knowledge to tackle advanced video generation models such as Sora and Seedance. Second, exploring efficient reasoning and logical distillation. By employing dynamic Chain-of-Thought truncation or distilling the complex reasoning capabilities of large models into lightweight architectures, we aim to overcome the latency induced by Long CoT and fulfill the strict real-time requirements of industrial deployment. Third, introducing Continual Learning mechanisms. This will enable the model to dynamically assimilate newly emerging algorithmic artifacts without catastrophic forgetting, ultimately maintaining long-term robustness in the open-world adversarial arms race.

\bibliographystyle{IEEEtran}
\bibliography{main}

\begin{thebibliography}{10}
\providecommand{\url}[1]{#1}
\csname url@samestyle\endcsname
\providecommand{\newblock}{\relax}
\providecommand{\bibinfo}[2]{#2}
\providecommand{\BIBentrySTDinterwordspacing}{\spaceskip=0pt\relax}
\providecommand{\BIBentryALTinterwordstretchfactor}{4}
\providecommand{\BIBentryALTinterwordspacing}{\spaceskip=\fontdimen2\font plus
\BIBentryALTinterwordstretchfactor\fontdimen3\font minus \fontdimen4\font\relax}
\providecommand{\BIBforeignlanguage}[2]{{%
\expandafter\ifx\csname l@#1\endcsname\relax
\typeout{** WARNING: IEEEtran.bst: No hyphenation pattern has been}%
\typeout{** loaded for the language `#1'. Using the pattern for}%
\typeout{** the default language instead.}%
\else
\language=\csname l@#1\endcsname
\fi
#2}}
\providecommand{\BIBdecl}{\relax}
\BIBdecl

\bibitem{goodfellow2014generative}
I.~J. Goodfellow, J.~Pouget-Abadie, M.~Mirza, B.~Xu, D.~Warde-Farley, S.~Ozair, A.~Courville, and Y.~Bengio, ``Generative adversarial nets,'' \emph{Advances in neural information processing systems}, vol.~27, 2014.

\bibitem{ho2020denoisingdiffusionprobabilisticmodels}
\BIBentryALTinterwordspacing
J.~Ho, A.~Jain, and P.~Abbeel, ``Denoising diffusion probabilistic models,'' 2020. [Online]. Available: \url{https://arxiv.org/abs/2006.11239}
\BIBentrySTDinterwordspacing

\bibitem{betker2023improving}
J.~Betker, G.~Goh, L.~Jing, T.~Brooks, J.~Wang, L.~Li, L.~Ouyang, J.~Zhuang, J.~Lee, Y.~Guo \emph{et~al.}, ``Improving image generation with better captions,'' \emph{Computer Science. https://cdn. openai. com/papers/dall-e-3. pdf}, vol.~2, no.~3, p.~8, 2023.

\bibitem{imageteam2025zimageefficientimagegeneration}
\BIBentryALTinterwordspacing
I.~Team, H.~Cai, S.~Cao, R.~Du, P.~Gao, S.~Hoi, Z.~Hou, S.~Huang, D.~Jiang, X.~Jin, L.~Li, Z.~Li, Z.-Y. Li, D.~Liu, D.~Liu, J.~Shi, Q.~Wu, F.~Yu, C.~Zhang, S.~Zhang, and S.~Zhou, ``Z-image: An efficient image generation foundation model with single-stream diffusion transformer,'' 2025. [Online]. Available: \url{https://arxiv.org/abs/2511.22699}
\BIBentrySTDinterwordspacing

\bibitem{dhariwal2021diffusion}
P.~Dhariwal and A.~Nichol, ``Diffusion models beat gans on image synthesis,'' \emph{Advances in neural information processing systems}, vol.~34, pp. 8780--8794, 2021.

\bibitem{rombach2022high}
R.~Rombach, A.~Blattmann, D.~Lorenz, P.~Esser, and B.~Ommer, ``High-resolution image synthesis with latent diffusion models,'' in \emph{Proceedings of the IEEE/CVF conference on computer vision and pattern recognition}, 2022, pp. 10\,684--10\,695.

\bibitem{croitoru2023diffusion}
F.-A. Croitoru, V.~Hondru, R.~T. Ionescu, and M.~Shah, ``Diffusion models in vision: A survey,'' \emph{IEEE transactions on pattern analysis and machine intelligence}, vol.~45, no.~9, pp. 10\,850--10\,869, 2023.

\bibitem{ye2025realgen}
J.~Ye, L.~Zhu, Y.~Guo, D.~Jiang, Z.~Huang, Y.~Zhang, Z.~Yan, H.~Fu, C.~He, and W.~Li, ``Realgen: Photorealistic text-to-image generation via detector-guided rewards,'' \emph{arXiv preprint arXiv:2512.00473}, 2025.

\bibitem{liu2024mmfakebench}
X.~Liu, Z.~Li, P.~Li, H.~Huang, S.~Xia, X.~Cui, L.~Huang, W.~Deng, and Z.~He, ``Mmfakebench: A mixed-source multimodal misinformation detection benchmark for lvlms,'' \emph{arXiv preprint arXiv:2406.08772}, 2024.

\bibitem{mustak2023deepfakes}
M.~Mustak, J.~Salminen, M.~M{\"a}ntym{\"a}ki, A.~Rahman, and Y.~K. Dwivedi, ``Deepfakes: Deceptions, mitigations, and opportunities,'' \emph{Journal of Business Research}, vol. 154, p. 113368, 2023.

\bibitem{koutlis2024leveraging}
C.~Koutlis and S.~Papadopoulos, ``Leveraging representations from intermediate encoder-blocks for synthetic image detection,'' in \emph{European Conference on computer vision}.\hskip 1em plus 0.5em minus 0.4em\relax Springer, 2024, pp. 394--411.

\bibitem{lin2025seeing}
K.~Lin, Z.~Yan, R.~Chen, J.~Ye, K.-Y. Zhang, Y.~Zhou, P.~Jin, B.~Li, T.~Yao, and S.~Ding, ``Seeing before reasoning: A unified framework for generalizable and explainable fake image detection,'' \emph{arXiv preprint arXiv:2509.25502}, 2025.

\bibitem{jiang2025ivy}
C.~Jiang, W.~Dong, Z.~Zhang, C.~Si, F.~Yu, W.~Peng, X.~Yuan, Y.~Bi, M.~Zhao, Z.~Zhou \emph{et~al.}, ``Ivy-fake: A unified explainable framework and benchmark for image and video aigc detection,'' \emph{arXiv preprint arXiv:2506.00979}, 2025.

\bibitem{guo2025omniaid}
Y.~Guo, J.~Ye, C.~Zhang, H.~Kang, H.~Fu, C.~He, and W.~Li, ``Omniaid: Decoupling semantic and artifacts for universal ai-generated image detection in the wild,'' \emph{arXiv preprint arXiv:2511.08423}, 2025.

\bibitem{he2021forgerynet}
Y.~He, B.~Gan, S.~Chen, Y.~Zhou, G.~Yin, L.~Song, L.~Sheng, J.~Shao, and Z.~Liu, ``Forgerynet: A versatile benchmark for comprehensive forgery analysis,'' in \emph{Proceedings of the IEEE/CVF conference on computer vision and pattern recognition}, 2021, pp. 4360--4369.

\bibitem{zhu2023genimage}
M.~Zhu, H.~Chen, Q.~Yan, X.~Huang, G.~Lin, W.~Li, Z.~Tu, H.~Hu, J.~Hu, and Y.~Wang, ``Genimage: A million-scale benchmark for detecting ai-generated image,'' \emph{Advances in neural information processing systems}, vol.~36, pp. 77\,771--77\,782, 2023.

\bibitem{chen2024drct}
B.~Chen, J.~Zeng, J.~Yang, and R.~Yang, ``Drct: Diffusion reconstruction contrastive training towards universal detection of diffusion generated images,'' in \emph{Forty-first International Conference on Machine Learning}, 2024.

\bibitem{hong2025wildfake}
Y.~Hong, J.~Feng, H.~Chen, J.~Lan, H.~Zhu, W.~Wang, and J.~Zhang, ``Wildfake: A large-scale and hierarchical dataset for ai-generated images detection,'' in \emph{Proceedings of the AAAI Conference on Artificial Intelligence}, vol.~39, no.~4, 2025, pp. 3500--3508.

\bibitem{jeong2022frepgan}
Y.~Jeong, D.~Kim, Y.~Ro, and J.~Choi, ``Frepgan: robust deepfake detection using frequency-level perturbations,'' in \emph{Proceedings of the AAAI conference on artificial intelligence}, vol.~36, no.~1, 2022, pp. 1060--1068.

\bibitem{chen2024single}
J.~Chen, J.~Yao, and L.~Niu, ``A single simple patch is all you need for ai-generated image detection,'' \emph{arXiv preprint arXiv:2402.01123}, 2024.

\bibitem{li2025fakescopelargemultimodalexpert}
\BIBentryALTinterwordspacing
Y.~Li, Y.~Tian, Y.~Huang, W.~Lu, S.~Wang, W.~Lin, and A.~Rocha, ``Fakescope: Large multimodal expert model for transparent ai-generated image forensics,'' 2025. [Online]. Available: \url{https://arxiv.org/abs/2503.24267}
\BIBentrySTDinterwordspacing

\bibitem{kang2025legion}
H.~Kang \emph{et~al.}, ``Legion: Learning to ground and explain for synthetic image detection,'' in \emph{Proceedings of the IEEE/CVF International Conference on Computer Vision (ICCV)}, 2025.

\bibitem{jia2024can}
S.~Jia, R.~Lyu, K.~Zhao, Y.~Chen, Z.~Yan, Y.~Ju, C.~Hu, X.~Li, B.~Wu, and S.~Lyu, ``Can chatgpt detect deepfakes? a study of using multimodal large language models for media forensics,'' in \emph{Proceedings of the IEEE/CVF Conference on Computer Vision and Pattern Recognition}, 2024, pp. 4324--4333.

\bibitem{ye2024loki}
J.~Ye, B.~Zhou, Z.~Huang, J.~Zhang, T.~Bai, H.~Kang, J.~He, H.~Lin, Z.~Wang, T.~Wu \emph{et~al.}, ``Loki: A comprehensive synthetic data detection benchmark using large multimodal models,'' \emph{arXiv preprint arXiv:2410.09732}, 2024.

\bibitem{li2025fakebench}
Y.~Li, X.~Liu, X.~Wang, B.~S. Lee, S.~Wang, A.~Rocha, and W.~Lin, ``Fakebench: Probing explainable fake image detection via large multimodal models,'' \emph{IEEE Transactions on Information Forensics and Security}, 2025.

\bibitem{chen2024x2}
Y.~Chen, Z.~Yan, G.~Cheng, K.~Zhao, S.~Lyu, and B.~Wu, ``X2-dfd: A framework for explainable and extendable deepfake detection,'' \emph{arXiv preprint arXiv:2410.06126}, 2024.

\bibitem{zhou2025aigi}
Z.~Zhou, Y.~Luo, Y.~Wu, K.~Sun, J.~Ji, K.~Yan, S.~Ding, X.~Sun, Y.~Wu, and R.~Ji, ``Aigi-holmes: Towards explainable and generalizable ai-generated image detection via multimodal large language models,'' in \emph{Proceedings of the IEEE/CVF International Conference on Computer Vision}, 2025, pp. 18\,746--18\,758.

\bibitem{wen2025spot}
S.~Wen, J.~Ye, P.~Feng, H.~Kang, Z.~Wen, Y.~Chen, J.~Wu, W.~Wu, C.~He, and W.~Li, ``Spot the fake: Large multimodal model-based synthetic image detection with artifact explanation,'' \emph{arXiv preprint arXiv:2503.14905}, 2025.

\bibitem{wang2501critique}
Y.~Wang, X.~Yue, and W.~Chen, ``Critique fine-tuning: Learning to critique is more effective than learning to imitate, 2025,'' \emph{URL https://arxiv. org/abs/2501.17703}.

\bibitem{wei2022chain}
J.~Wei, X.~Wang, D.~Schuurmans, M.~Bosma, F.~Xia, E.~Chi, Q.~V. Le, D.~Zhou \emph{et~al.}, ``Chain-of-thought prompting elicits reasoning in large language models,'' \emph{Advances in neural information processing systems}, vol.~35, pp. 24\,824--24\,837, 2022.

\bibitem{ye2025echo}
J.~Ye, D.~Jiang, Z.~Wang, L.~Zhu, Z.~Hu, Z.~Huang, J.~He, Z.~Yan, J.~Yu, H.~Li \emph{et~al.}, ``Echo-4o: Harnessing the power of gpt-4o synthetic images for improved image generation,'' \emph{arXiv preprint arXiv:2508.09987}, 2025.

\bibitem{yan2025gpt}
Z.~Yan, J.~Ye, W.~Li, Z.~Huang, S.~Yuan, X.~He, K.~Lin, J.~He, C.~He, and L.~Yuan, ``Gpt-imgeval: A comprehensive benchmark for diagnosing gpt4o in image generation,'' \emph{arXiv preprint arXiv:2504.02782}, 2025.

\bibitem{he2026mind}
J.~He, J.~Ye, Z.~Huang, D.~Jiang, C.~Zhang, L.~Zhu, R.~Zhang, X.~Zhang, and W.~Li, ``Mind-brush: Integrating agentic cognitive search and reasoning into image generation,'' \emph{arXiv preprint arXiv:2602.01756}, 2026.

\bibitem{wang2020cnn}
S.-Y. Wang, O.~Wang, R.~Zhang, A.~Owens, and A.~A. Efros, ``Cnn-generated images are surprisingly easy to spot... for now,'' in \emph{Proceedings of the IEEE/CVF conference on computer vision and pattern recognition}, 2020, pp. 8695--8704.

\bibitem{ojha2023towards}
U.~Ojha, Y.~Li, and Y.~J. Lee, ``Towards universal fake image detectors that generalize across generative models,'' in \emph{Proceedings of the IEEE/CVF conference on computer vision and pattern recognition}, 2023, pp. 24\,480--24\,489.

\bibitem{bi2023detecting}
X.~Bi, B.~Liu, F.~Yang, B.~Xiao, W.~Li, G.~Huang, and P.~C. Cosman, ``Detecting generated images by real images only,'' \emph{arXiv preprint arXiv:2311.00962}, 2023.

\bibitem{goodfellow2020generative}
I.~Goodfellow, J.~Pouget-Abadie, M.~Mirza, B.~Xu, D.~Warde-Farley, S.~Ozair, A.~Courville, and Y.~Bengio, ``Generative adversarial networks,'' \emph{Communications of the ACM}, vol.~63, no.~11, pp. 139--144, 2020.

\bibitem{zhong2023rich}
N.~Zhong, Y.~Xu, Z.~Qian, and X.~Zhang, ``Rich and poor texture contrast: A simple yet effective approach for ai-generated image detection,'' \emph{arXiv preprint arXiv:2311.12397}, vol.~3, no.~6, p.~1, 2023.

\bibitem{ciftci2020fakecatcher}
U.~A. Ciftci, I.~Demir, and L.~Yin, ``Fakecatcher: Detection of synthetic portrait videos using biological signals,'' \emph{IEEE transactions on pattern analysis and machine intelligence}, 2020.

\bibitem{wolter2022wavelet}
M.~Wolter, F.~Blanke, R.~Heese, and J.~Garcke, ``Wavelet-packets for deepfake image analysis and detection,'' \emph{Machine Learning}, vol. 111, no.~11, pp. 4295--4327, 2022.

\bibitem{tan2024rethinking}
C.~Tan, Y.~Zhao, S.~Wei, G.~Gu, P.~Liu, and Y.~Wei, ``Rethinking the up-sampling operations in cnn-based generative network for generalizable deepfake detection,'' in \emph{Proceedings of the IEEE/CVF conference on computer vision and pattern recognition}, 2024, pp. 28\,130--28\,139.

\bibitem{gupta2024visual}
A.~S. Gupta, K.~P. Shreneter, and S.~Sehgal, ``Visual veracity: Advancing ai-generated image detection with convolutional neural networks,'' \emph{2024 11th International Conference on Reliability, Infocom Technologies and Optimization (Trends and Future Directions) (ICRITO)}, 2024.

\bibitem{wani2025comparative}
P.~Wani, S.~Chavan, S.~Paithankar, D.~Ghusse, and S.~Barve, ``Comparative analyais of cnn architectures for deep fake detection,'' \emph{2025 3rd International Conference on Intelligent Systems, Advanced Computing and Communication (ISACC)}, 2025.

\bibitem{lađević2024detection}
A.~L. Lađević, T.~Kramberger, R.~Kramberger, and D.~Vlahek, ``Detection of ai-generated synthetic images with a lightweight cnn,'' \emph{Applied Informatics}, 2024.

\bibitem{hossain2023advancing}
M.~Z. Hossain, F.~U. Zaman, and M.~R. Islam, ``Advancing ai-generated image detection: Enhanced accuracy through cnn and vision transformer models with explainable ai insights,'' \emph{2023 26th International Conference on Computer and Information Technology (ICCIT)}, 2023.

\bibitem{chang2023antifakeprompt}
Y.-M. Chang, C.~Yeh, W.-C. Chiu, and N.~Yu, ``Antifakeprompt: Prompt-tuned vision-language models are fake image detectors,'' \emph{arXiv preprint arXiv:2310.17419}, 2023.

\bibitem{liu2022fad}
X.~Liu, J.~Liu, P.~Guo, D.~Tuo, S.~Tian, and Y.~Jiang, ``Fad-net: Fake images detection and generalization based on frequency domain transformation,'' in \emph{2022 15th International Congress on Image and Signal Processing, BioMedical Engineering and Informatics (CISP-BMEI)}.\hskip 1em plus 0.5em minus 0.4em\relax IEEE, 2022, pp. 1--7.

\bibitem{tan2024frequency}
C.~Tan, Y.~Zhao, S.~Wei, G.~Gu, P.~Liu, and Y.~Wei, ``Frequency-aware deepfake detection: Improving generalizability through frequency space domain learning,'' in \emph{Proceedings of the AAAI Conference on Artificial Intelligence}, vol.~38, no.~5, 2024, pp. 5052--5060.

\bibitem{wang2023dynamic}
Y.~Wang, K.~Yu, C.~Chen, X.~Hu, and S.~Peng, ``Dynamic graph learning with content-guided spatial-frequency relation reasoning for deepfake detection,'' in \emph{Proceedings of the IEEE/CVF conference on computer vision and pattern recognition}, 2023, pp. 7278--7287.

\bibitem{gpt4o}
OpenAI, ``Gpt-4o,'' \url{https://openai.com/index/introducing-4o-image-generation}, 2025.

\bibitem{google2025gemini}
\BIBentryALTinterwordspacing
G.~DeepMind, ``Gemini 2.5 pro,'' Accessed: 2025-11-11, 2025. [Online]. Available: \url{https://cloud.google.com/vertex-ai/generative-ai/docs/models/gemini/2-5-pro}
\BIBentrySTDinterwordspacing

\bibitem{liu2023visual}
H.~Liu, C.~Li, Q.~Wu, and Y.~J. Lee, ``Visual instruction tuning,'' \emph{Advances in neural information processing systems}, vol.~36, pp. 34\,892--34\,916, 2023.

\bibitem{bai2025qwen25vltechnicalreport}
\BIBentryALTinterwordspacing
S.~Bai, K.~Chen, X.~Liu, J.~Wang, W.~Ge, S.~Song, K.~Dang, P.~Wang, S.~Wang, J.~Tang, H.~Zhong, Y.~Zhu, M.~Yang, Z.~Li, J.~Wan, P.~Wang, W.~Ding, Z.~Fu, Y.~Xu, J.~Ye, X.~Zhang, T.~Xie, Z.~Cheng, H.~Zhang, Z.~Yang, H.~Xu, and J.~Lin, ``Qwen2.5-vl technical report,'' 2025. [Online]. Available: \url{https://arxiv.org/abs/2502.13923}
\BIBentrySTDinterwordspacing

\bibitem{wu2023q}
H.~Wu, Z.~Zhang, E.~Zhang, C.~Chen, L.~Liao, A.~Wang, C.~Li, W.~Sun, Q.~Yan, G.~Zhai \emph{et~al.}, ``Q-bench: A benchmark for general-purpose foundation models on low-level vision,'' \emph{arXiv preprint arXiv:2309.14181}, 2023.

\bibitem{zhang2024visually}
Y.~Zhang, A.~Unell, X.~Wang, D.~Ghosh, Y.~Su, L.~Schmidt, and S.~Yeung-Levy, ``Why are visually-grounded language models bad at image classification?'' \emph{Advances in Neural Information Processing Systems}, vol.~37, pp. 51\,727--51\,753, 2024.

\bibitem{xu2024fakeshield}
Z.~Xu, X.~Zhang, R.~Li, Z.~Tang, Q.~Huang, and J.~Zhang, ``Fakeshield: Explainable image forgery detection and localization via multi-modal large language models,'' \emph{arXiv preprint arXiv:2410.02761}, 2024.

\bibitem{zhang2024common}
Y.~Zhang, B.~Colman, X.~Guo, A.~Shahriyari, and G.~Bharaj, ``Common sense reasoning for deepfake detection,'' in \emph{European conference on computer vision}.\hskip 1em plus 0.5em minus 0.4em\relax Springer, 2024, pp. 399--415.

\bibitem{huang2024ffaa}
Z.~Huang, B.~Xia, Z.~Lin, Z.~Mou, W.~Yang, and J.~Jia, ``Ffaa: Multimodal large language model based explainable open-world face forgery analysis assistant,'' \emph{arXiv preprint arXiv:2408.10072}, 2024.

\bibitem{wen2025busterx++}
H.~Wen, T.~Li, Z.~Huang, Y.~He, and G.~Cheng, ``Busterx++: Towards unified cross-modal ai-generated content detection and explanation with mllm,'' \emph{arXiv preprint arXiv:2507.14632}, 2025.

\bibitem{li2025skyra}
Y.~Li, W.~Zheng, Y.~Zhang, R.~Sun, Y.~Zheng, L.~Chen, J.~Zhou, and J.~Lu, ``Skyra: Ai-generated video detection via grounded artifact reasoning,'' \emph{arXiv preprint arXiv:2512.15693}, 2025.

\bibitem{liu2024forgerygpt}
J.~Liu, F.~Zhang, J.~Zhu, E.~Sun, Q.~Zhang, and Z.-J. Zha, ``Forgerygpt: Multimodal large language model for explainable image forgery detection and localization,'' \emph{arXiv preprint arXiv:2410.10238}, 2024.

\bibitem{huang2025sofake}
\BIBentryALTinterwordspacing
Z.~Huang, T.~Li, X.~Li, H.~Wen, Y.~He, J.~Zhang, H.~Fei, X.~Yang, X.~Huang, B.~Peng, and G.~Cheng, ``So-fake: Benchmarking and explaining social media image forgery detection,'' 2025. [Online]. Available: \url{https://arxiv.org/abs/2505.18660}
\BIBentrySTDinterwordspacing

\bibitem{huang2025sida}
Z.~Huang, J.~Hu, X.~Li, Y.~He, X.~Zhao, B.~Peng, B.~Wu, X.~Huang, and G.~Cheng, ``Sida: Social media image deepfake detection, localization and explanation with large multimodal model,'' in \emph{Proceedings of the Computer Vision and Pattern Recognition Conference}, 2025, pp. 28\,831--28\,841.

\bibitem{Kuznetsova_2020}
\BIBentryALTinterwordspacing
A.~Kuznetsova, H.~Rom, N.~Alldrin, J.~Uijlings, I.~Krasin, J.~Pont-Tuset, S.~Kamali, S.~Popov, M.~Malloci, A.~Kolesnikov, T.~Duerig, and V.~Ferrari, ``The open images dataset v4: Unified image classification, object detection, and visual relationship detection at scale,'' \emph{International Journal of Computer Vision}, vol. 128, no.~7, p. 1956–1981, Mar. 2020. [Online]. Available: \url{http://dx.doi.org/10.1007/s11263-020-01316-z}
\BIBentrySTDinterwordspacing

\bibitem{kwon2023efficient}
W.~Kwon, Z.~Li, S.~Zhuang, Y.~Sheng, L.~Zheng, C.~H. Yu, J.~Gonzalez, H.~Zhang, and I.~Stoica, ``Efficient memory management for large language model serving with pagedattention,'' in \emph{Proceedings of the 29th symposium on operating systems principles}, 2023, pp. 611--626.

\bibitem{zheng2024sglang}
L.~Zheng, L.~Yin, Z.~Xie, C.~Sun, J.~Huang, C.~H. Yu, S.~Cao, C.~Kozyrakis, I.~Stoica, J.~E. Gonzalez \emph{et~al.}, ``Sglang: Efficient execution of structured language model programs,'' \emph{Advances in neural information processing systems}, vol.~37, pp. 62\,557--62\,583, 2024.

\bibitem{guo2025deepseek}
D.~Guo, D.~Yang, H.~Zhang, J.~Song, P.~Wang, Q.~Zhu, R.~Xu, R.~Zhang, S.~Ma, X.~Bi \emph{et~al.}, ``Deepseek-r1: Incentivizing reasoning capability in llms via reinforcement learning,'' \emph{arXiv preprint arXiv:2501.12948}, 2025.

\bibitem{fan2025sophiavl}
K.~Fan, K.~Feng, H.~Lyu, D.~Zhou, and X.~Yue, ``Sophiavl-r1: Reinforcing mllms reasoning with thinking reward,'' \emph{arXiv preprint arXiv:2505.17018}, 2025.

\bibitem{corvi2023detection}
R.~Corvi, D.~Cozzolino, G.~Zingarini, G.~Poggi, K.~Nagano, and L.~Verdoliva, ``On the detection of synthetic images generated by diffusion models,'' in \emph{ICASSP 2023-2023 IEEE International Conference on Acoustics, Speech and Signal Processing (ICASSP)}.\hskip 1em plus 0.5em minus 0.4em\relax IEEE, 2023, pp. 1--5.

\bibitem{yan2024effort}
Z.~Yan, J.~Wang, Z.~Wang, P.~Jin, K.-Y. Zhang, S.~Chen, T.~Yao, S.~Ding, B.~Wu, and L.~Yuan, ``Effort: Efficient orthogonal modeling for generalizable ai-generated image detection,'' \emph{arXiv preprint arXiv:2411.15633}, 2024.

\bibitem{yan2024sanity}
S.~Yan, O.~Li, J.~Cai, Y.~Hao, X.~Jiang, Y.~Hu, and W.~Xie, ``A sanity check for ai-generated image detection,'' \emph{arXiv preprint arXiv:2406.19435}, 2024.

\bibitem{openai_gpt5_system_card_2025}
{OpenAI}, ``Gpt-5 system card,'' \url{https://cdn.openai.com/gpt-5-system-card.pdf}, Aug. 2025.

\bibitem{wu2024deepseekvl2mixtureofexpertsvisionlanguagemodels}
\BIBentryALTinterwordspacing
Z.~Wu, X.~Chen, Z.~Pan, X.~Liu, W.~Liu, D.~Dai, H.~Gao, Y.~Ma, C.~Wu, B.~Wang, Z.~Xie, Y.~Wu, K.~Hu, J.~Wang, Y.~Sun, Y.~Li, Y.~Piao, K.~Guan, A.~Liu, X.~Xie, Y.~You, K.~Dong, X.~Yu, H.~Zhang, L.~Zhao, Y.~Wang, and C.~Ruan, ``Deepseek-vl2: Mixture-of-experts vision-language models for advanced multimodal understanding,'' 2024. [Online]. Available: \url{https://arxiv.org/abs/2412.10302}
\BIBentrySTDinterwordspacing

\bibitem{zhu2025internvl3exploringadvancedtraining}
\BIBentryALTinterwordspacing
J.~Zhu, W.~Wang, Z.~Chen, Z.~Liu, S.~Ye, L.~Gu, H.~Tian, Y.~Duan, W.~Su, J.~Shao, Z.~Gao, E.~Cui, X.~Wang, Y.~Cao, Y.~Liu, X.~Wei, H.~Zhang, H.~Wang, W.~Xu, H.~Li, J.~Wang, N.~Deng, S.~Li, Y.~He, T.~Jiang, J.~Luo, Y.~Wang, C.~He, B.~Shi, X.~Zhang, W.~Shao, J.~He, Y.~Xiong, W.~Qu, P.~Sun, P.~Jiao, H.~Lv, L.~Wu, K.~Zhang, H.~Deng, J.~Ge, K.~Chen, L.~Wang, M.~Dou, L.~Lu, X.~Zhu, T.~Lu, D.~Lin, Y.~Qiao, J.~Dai, and W.~Wang, ``Internvl3: Exploring advanced training and test-time recipes for open-source multimodal models,'' 2025. [Online]. Available: \url{https://arxiv.org/abs/2504.10479}
\BIBentrySTDinterwordspacing

\bibitem{liu2020global}
Z.~Liu, X.~Qi, and P.~H. Torr, ``Global texture enhancement for fake face detection in the wild,'' in \emph{Proceedings of the IEEE/CVF conference on computer vision and pattern recognition}, 2020, pp. 8060--8069.

\bibitem{ju2022fusing}
Y.~Ju, S.~Jia, L.~Ke, H.~Xue, K.~Nagano, and S.~Lyu, ``Fusing global and local features for generalized ai-synthesized image detection,'' in \emph{2022 IEEE International Conference on Image Processing (ICIP)}.\hskip 1em plus 0.5em minus 0.4em\relax IEEE, 2022, pp. 3465--3469.

\end{thebibliography}

\vfill

\end{document}